\documentclass[atmp,a4paper]{ipart_v2}

\Vol{30}
\Issue{1}
\Year{2026}
\Doi{10.4310/ATMP.260413005111}
\Pubonline{April 29, 2026}
\firstpage{189}

\usepackage[english]{babel}
\usepackage[utf8]{inputenc}

\usepackage{graphicx}
\usepackage{amssymb}
\usepackage{amsmath}
\usepackage{amsfonts}
\usepackage{amsthm}
\usepackage{color}
\usepackage[mathscr]{eucal}
\usepackage{enumerate}
\usepackage{algorithm}
\usepackage{algorithmic}
\usepackage[skip=10pt]{caption} 

\newtheorem{Def}{Definition}[section]

\newcommand{\beq}{\begin{equation}}
\newcommand{\eeq}{\end{equation}}
\newcommand{\Proof}{\begin{proof}}
\newcommand{\QED}{\end{proof} \noindent}

\newtheorem{proposition}[Def]{Proposition}

\begin{document} 

\title[CayleyPy RL]{CayleyPy RL: Pathfinding and reinforcement learning on Cayley graphs}
\author[A. Soibelman et al.]{
Alexander Chervov\textsuperscript{*},
Mark Obozov\textsuperscript{*},
Alexander Soibelman,
Sergey Lytkin,
Igor Kiselev,
Sergei Fironov,
Andrey Lukyanenko,
Antonina Dolgorukova,
Andrey Ogurtsov,
Fedor Petrov,
Stanislav Krymskii,
Mikhail Evseev,
Lilia Grunwald,
Denis Gorodkov,
Grigorii Antiufeev,
Gordey Verbii,
Vladislav Zamkovoy,
Liuda Cheldieva,
Ivan Koltsov,
Arseniy Sychev,
Anton Eliseev,
Sergei Nikolenko,
Nursultan Nanrynbaev,
Rustem Turtaev,
Nikita Rokotyan,
Sviatoslav Kovalev,
Alexei Rozanov,
Veniamin Nelin,
Sergei Ermilov,
Lidia Shishina,
Dariya Mamayeva,
Antonina Korolkova,
Kirill Khoruzhii,
and Alexey Romanov
}

\footnotetext[0]{\textsuperscript{*}These authors contributed equally to this work.}

\begin{abstract}
This paper is the second in a series of studies on developing efficient artificial intelligence-based approaches to pathfinding on extremely large graphs (e.g., $10^{70}$ nodes) focusing on Cayley graphs and mathematical applications. The open-source CayleyPy project is a central component of our research. The present paper proposes a novel combination of a reinforcement learning approach with a more direct diffusion distance approach from the first paper. Our analysis includes benchmarking various choices for the key building blocks of the approach: neural network architectures, random walk generators, and beam search pathfinding. We compared these methods against the classical computer algebra system GAP, demonstrating that they "overcome the GAP" for the considered examples.
We examine the Cayley graph of the symmetric group with cyclic shift and transposition generators as a particular mathematical application. We strongly support the OEIS-A186783 conjecture that the diameter equals n(n-1)/2 by machine learning and mathematical methods. We identify the conjectured longest element and generate its decomposition of the desired length. We prove a diameter lower bound of $n(n-1)/2-n/2-1$ and an upper bound of $n(n-1)/2+3n$ by presenting the algorithm with given complexity. We also present several conjectures motivated by numerical experiments, including observations on the central limit phenomenon (with growth approximated by the Gumbel distribution), the uniform distribution for the spectrum of the graph, and a numerical study of sorting networks.
We create challenges on the Kaggle platform to stimulate crowdsourcing activity and invite contributions to improve and benchmark approaches to Cayley graph pathfinding and other tasks.
\end{abstract}


\maketitle 
\tableofcontents

\section{Introduction}

\subsection{Broader context}
Deep learning has revolutionized various fields of research and was recognized with several Nobel Prizes in 2024.  
In recent years, machine learning has been emerging as "a tool in theoretical science"~\cite{douglas2022machine}, leading to several noteworthy applications to mathematical problems:~\cite{lample2019deep,davies2021advancing, bao2021polytopes, 
romera2024mathematical,
coates2024machine,alfarano2024global, charton2024patternboost,shehper2024makes,swirszcz2025advancing}.  

The paper presents advancements in applying artificial intelligence methods to the mathematical problem of Cayley graph pathfinding, equivalent to decomposing a group element into a product of generators. It is the second work in a series devoted to the CayleyPy project to develop efficient machine learning-based approaches and an open-source library to deal with extra-large graphs.
(e.g. $10^{70}$) with a focus on finite Cayley graphs and mathematical applications.

From a broader perspective, pathfinding is a specific case of the planning problem, where one must determine a sequence of actions to transition between two given states. Similar challenges arise in robotics and games like chess or Go, where the objective is to plan moves that lead from the initial position to a winning position. Mathematically, such problems are modeled as pathfinding on graphs (state transition graphs), where nodes represent possible states, and edges correspond to state transitions based on actions or moves. The planning task then reduces to finding a path from a given initial node to one or more target nodes. The breakthrough works AlphaGo and AlphaZero~\cite{silver2016mastering}, 
\cite{silver2017mastering} by Google DeepMind have demonstrated that machine learning can significantly outperform previously known methods. They serve as both a prototype and an inspiration for many developments, including ours. The method consists of two main components: a neural network trained to suggest moves from any given state and a graph pathfinding algorithm that helps to find a path even when the neural network's predictions are not sufficiently precise. The core idea is similar across most approaches, but achieving optimal performance for each task requires analyzing and benchmarking various neural networks and graph pathfinding methods. 

\begin{table}[ht!]
    \centering
    \begin{tabular}{|c|c|c|}
        \hline
        Graph  & Group  & RL  \\
        \hline
        vertex / node & permutation / element & state \\
        edge & generator & action \\
        path & word  & trajectory \\
        shortest path & shortest word  & maximal return \\
        diameter & longest element & ?\\
        \hline
    \end{tabular}
    \captionsetup{skip=10pt} 
    \caption{Comparison of graph theory, algebraic and reinforcement learning terminology.}
    \label{tab:terms}
\end{table}

One of the goals of this work is to apply these methods to a particular class of Cayley graphs. As in our previous paper~\cite{chervov2025machinelearningapproachbeats}, we choose a set of \textit{generators} $a_1^{\pm 1},\ldots, a_k^{\pm 1}\in S_n$. All finite products (words) of these elements form a subgroup $G \subset S_n$ which we identify with its Cayley graph. Due to vertex-transitivity of Cayley graphs, the pathfinding problem can be reduced to searching a path from some element $g\in G$ to the identity permutation $e$; this is the same as finding a decomposition of $g$ into a word $\prod\limits_{j=1}^m a_{i_j}^{\pm 1}$, $1 \leq i_j \leq k$. The problem of finding the shortest path (or word) can also be reformulated in terms of a generic reinforcement learning approach discussed in Section~\ref{Bellman_section}. A set of dictionary between RL, graphs, and group  is presented in Table~\ref{tab:terms}.

\subsection{ Present case study - LRX Cayley graph  } 

\begin{figure}[ht!]
 \centering
 \includegraphics[width=0.9\linewidth]{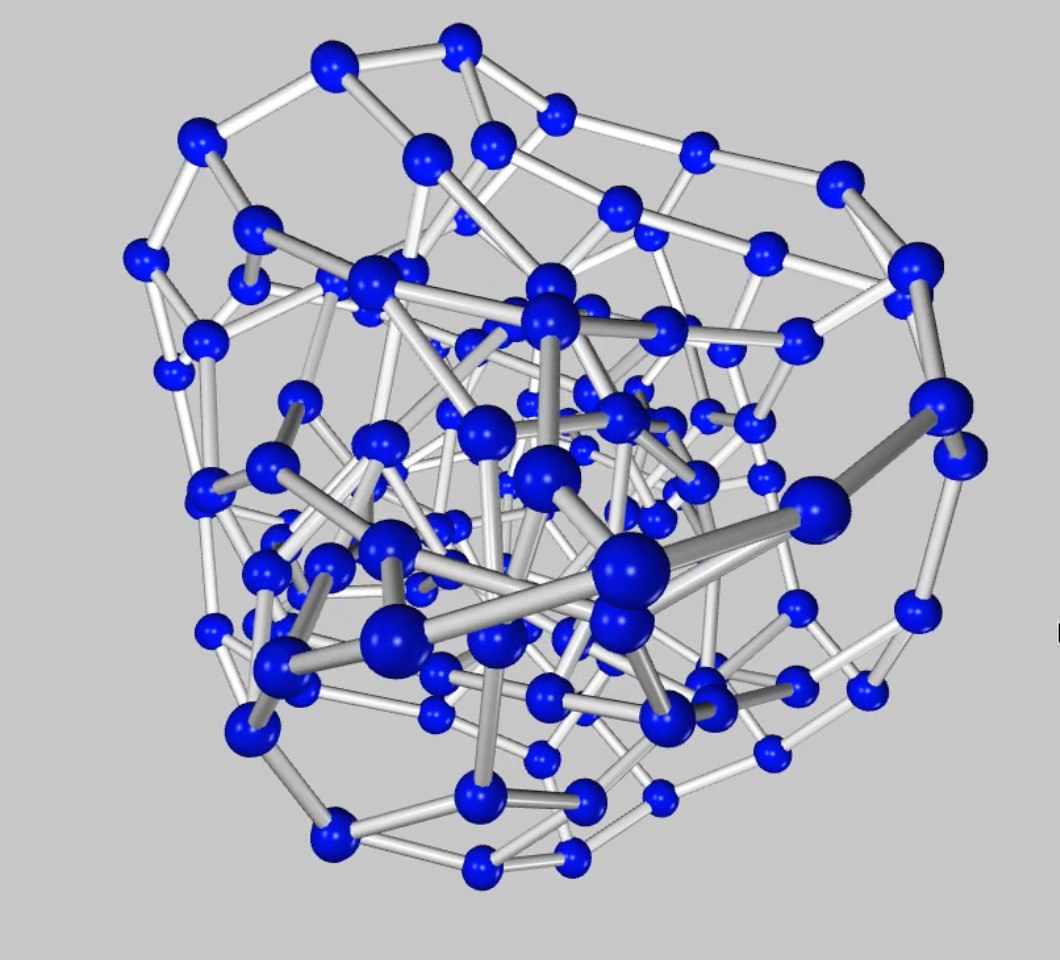}
 \caption{LRX Cayley graph for $S_5$}\label{fig:LRX1}
\end{figure}

The paper presents both AI contributions and mathematical results and conjectures. We use \textit{LRX Cayley graphs} as primary benchmark in the subsequent discussion.

\begin{figure}[ht!]
   \centering
   \includegraphics[width=1.0\linewidth]{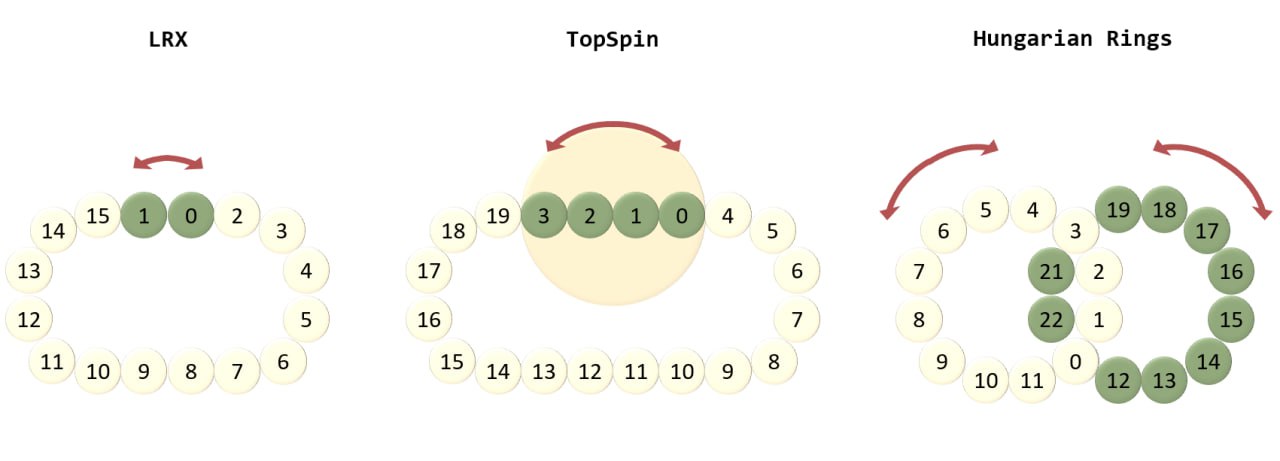}
   \caption{LRX and related mechanical puzzles}\label{fig:LRX_TopSpin_HR}
\end{figure}

{\bf LRX Cayley graphs.} 
For each $n\in\mathbb N$ the symmetric group $S_n$ can be generated by three permutations:

\begin{enumerate}
    \item 
    left cyclic shift $L = (1\ 2\ 3 \ \ldots\ n-1 \ 0)$;
    \item
    right cyclic shift $R = L^{-1} = (n-1\ 0\ 1 \ 2\  \ldots\ n-3 \ n-2)$;
    \item 
    transposition (eXchange) of the first two positions $X = X^{-1} = (0,1)$.
\end{enumerate}
These generators and the corresponding Cayley graph will be denoted LRX following \href{https://oeis.org/A186783}{OEIS-A186783}.
Visualization (Fig.~\ref{fig:LRX1}) of the LRX Cayley graph for $S_5$ reveals its complicated structure. 
LRX Cayley graph can be considered as a relative of the \href{https://www.sfu.ca/~jtmulhol/math302/puzzles-ot.html}{\textit{Oval Track}} (or \textit{TopSpin}) and \href{https://www.sfu.ca/~jtmulhol/math302/puzzles-hr.html}{\textit{Hungarian Rings}} puzzles (Fig.~\ref{fig:LRX_TopSpin_HR}).  

{\bf Conjectures on diameters of LRX and LX (V.M. Glushkov's 1968 problem).} 
It is conjectured \href{https://oeis.org/A186783}{OEIS-A186783} that diameters of LRX are $n(n-1)/2$,
that conjecture will be in the focus of the present paper. 
The case of two generators just and L and X (not inverse closed) goes back at least to 1968 influential paper by
``\href{https://en.wikipedia.org/wiki/Victor_Glushkov}{one of the founding fathers of Soviet cybernetics}'' V.\,M.~Glushkov 
\cite{glushkov1968completeness}, which has been much studied (survey: \cite{glukhovzubov1999lengths} pages 18-21),
but not resolved. The conjecture on the diameter in that case is presented in our subsequent paper: \cite{CayleyPyGrowth}.

Not exactly in the present scope, but worth mentioning: there was a long-standing problem resolved positively recently \cite{SawadaWilliams2019SigmaTau} of whether the directed graph generated by L and X (also denoted Sigma and Tau)
has a hamiltonian path.  D.Knuth assigned to that problem complexity 48 out of 50, it was first articulated in
\cite{NijenhuisWilf1975CombinatorialAlgorithms}.

\subsection{ AI contributions}~ 

{\bf Development of AI methodology.} We suggest a modified deep Q-learning (DQN) method for the case of the graph pathfinding task, which resolves the so-called sparse reward problem of the standard DQN approach for graph pathfinding. The proposal combines the DQN and the diffusion distance approach from~\cite{chervov2025machinelearningapproachbeats} and presents an analysis and benchmarks of these techniques, including tests of various neural network architectures such as MLP, CNN, and transformers. 
These methods have been shown to find paths for LRX Cayley graphs up to $n$ around 40 and with small modification (X-trick, which is specfic to LRX) even beyond 100.
We demonstrate that they "overcome the GAP": outperform the classical computer algebra system GAP on the same task, which can work only up to $n$ around 20, also producing much more optimal paths and working faster.

{\bf Efficient PyTorch implementation.} We devote significant attention to developing an original, highly optimized code. The current version is specific to Cayley graphs of permutation groups (matrix groups will be supported later). We propose many technical solutions for fast parallel processing of large amounts of permutations, including product computations, efficient hashing, and extracting unique states. The code works on both CPU and GPU without modification.

{\bf X-trick.} We report a curious finding: a single-line modification (which we call \textit{X-trick})  in the code of our main graph pathfinding module (beam search) dramatically improves performance. This enhancement increases the feasible pathfinding size
to $n$ around 100 (apparently even further with more powerful hardware). Unfortunately, this improvement is currently specific to the particular LRX Cayley graphs considered in this paper, but we hope it can be generalized. 
Moreover, we provide a mathematical counterpart of that finding - X-trick modified random walks are mixing much faster - see discussion below. 

\subsection{Mathematical contributions} 

We present several conjectures about LRX Cayley graphs which could leverage a more general understanding of the properties of that Cayley graph and its relatives.

{\bf Diameter conjecture and the longest element.} Surprisingly, according to \href{https://oeis.org/A186783}{OEIS-A186783}, it is an open conjecture that the diameter of the Cayley graph is $n(n-1)/2$ for $S_n$. 
By implementing an efficient brute-force breadth-first search algorithm, we traverse the full LRX graphs for $n\le 15$, confirming that the longest element has a length of $n(n-1)/2$, as predicted. Moreover, we observe that this element $\ell_n$ is always unique and follows a clear pattern

\begin{equation}
\label{eq:longest}
    \ell_n = (1 \ 0 \ n-1\ n-2\ \ldots\ 3\ 2) = (0,1)(n-1,2)(n-2,3)\ldots = R^2 e^{\leftarrow},
\end{equation}
where $e^{\leftarrow} = (n-1 \ n-2 \ \ldots \ 2 \ 1\ 0)$ is reversed identity permutation.
We expect~\eqref{eq:longest} holds for all $n$. Interestingly, the longest element $\ell_n$ is equal not to the reversed identity permutation itself but to the element obtained by two right shifts from it.

{\bf AI and crowd-sourcing to support the conjecture.} For the longest element $\ell_n$, we have launched various versions of our pipelines, and have consistently found that a decomposition is never shorter than $n(n-1)/2$. We organized a \href{https://www.kaggle.com/competitions/lrx-oeis-a-186783-brainstorm-math-conjecture/overview}{challenge} on the Kaggle platform suggesting to decompose $\ell_n$, $n=2, \ldots, 200$, into the shortest possible word.
Once again, no participant found a decomposition shorter than $n(n-1)/2$, thereby confirming the expectation.

{\bf Rigorous lower and upper bounds for the diameter.} We prove the lower bound $n(n-1)/2 - n/2-1$ by a combinatorial argument (improving $n(n-1)/3$~\cite{babai1989small}); later, we prove the upper bound $n(n-1)/2 +3n$ by the algorithm with such complexity (improving $\frac 32 n^2$~\cite{kuppili2020upper}). 
We also develop another algorithm that empirically shows better complexity $n(n-1)/2 + n/2$, but it lacks a rigorous proof for such an estimate.  
Finally, we present an explicit decomposition for the conjecturally longest element $\ell_n$ of the desired length $n(n-1)/2$. 

{\bf Central limit phenomena for growth.} 
Computations suggest the following conjecture: for large $n$, the growth function of LRX graphs follows a left-skewed \href{https://en.wikipedia.org/wiki/Gumbel_distribution}{Gumbel distribution}. This statement is in the vein of field-shaping works by P. Diaconis et. al.~\cite{diaconis1977spearman, diaconis1988metrics,chatterjee2017central}, who demonstrated that the growth of the neighbor transposition graph and many related statistics asymptotically follows the Gaussian normal distribution.
However we observe that skewness on the growth distribution for LRX does not tend to zero, thus Gaussian is not a choice, while Gumbel provides a good numerical fit. We also present a similar analysis for the Schreier coset graph for the action of the LRX generators on binary strings consisting of $n/2$ zeros and ones (for significantly larger range of $n$).

{\bf Random walks and mixing time.}
P. Diaconis and L. Saloff-Coste~\cite{diaconis1993comparison} obtained an important result that mixing time for LRX
is between $O(n^3)$ and $O(n^3log(n))$. We support their result numerically and complement it by considering two other types
of random walks: non-backtracking and X-trick modified. Both show faster mixing, however non-backtracking is only faster 
by a constant  factor, while X-trick leads to around $O(n^2)$ mixing. Which particially explains its effectivity in beam search.

{\bf Spectrum and sorting networks.}
We study numerically the spectrum of the LRX graph and observe a surprisingly uniform distribution of eigenvalues. Investigating statistics of the paths from $\ell_n$ to $e$ ("sorting networks"), we observe numerically that the pattern differs from the famous "sine curves" for Cayley graph of neighbor transpositions~\cite{angel2007random}.


{\bf In summary}, our analysis demonstrates that computational experiments, especially those involving AI, can be very helpful in advancing mathematical research on Cayley graphs. AI methods significantly outperform classical methods of computer algebra systems such as GAP, performing well for $S_n$ with $n$ around 40 and with a small addition of prior knowledge up to 90, while classical methods perform well only up to 20. 

\subsection{Review. Finite Cayley graphs: pathfinding techniques, results and open problems}

Let us briefly summarize existing techniques for Cayley graphs pathfinding, and provide more general context on Cayley graphs research.

{\bf Optimal pathfinding is typically NP-hard.} Finding the shortest paths on generic finite Cayley graphs is an NP-hard problem~\cite{even1981minimum} (even P-space complete \cite{jerrum1985complexity}). As is the case for many specific group families, such as $N\times N \times N$ Rubik's Cube groups~\cite{demaine2017solving} and others~\cite{bulteau2015pancake}.
It is hard to develop a practical optimal algorithm for particular groups, for instance, the first optimal algorithm for the standard $3\times 3 \times 3$ Rubik's cube has been proposed only in 1997~\cite{korf1997finding}. It required 82 Megabytes of precomputed data and 12-24 hours to solve a single cube (later improved~\footnote{M.Reid: https://www.cflmath.com/Rubik/optimal\_solver.html, H.Kociemba 2021: https://pypi.org/project/RubikOptimal/, Andrew Skalski:  https://github.com/Voltara/vcubem , \cite{Rokicki2014Diameter} }
to several cubes per second). Developing a practical optimal solvers for a $4\times 4 \times 4$ Rubik's cube is challenging open problem. 

{\bf Non-optimal pathfinding: Schreier-Sims algorithm, GAP implementation, non-effectivity for large groups.} 
The Schreier-Sims algorithm~\cite{sims1970computational} (and its randomized version \cite{knuth1991efficient}) is the method typically used for decomposing group elements into products of generators for permutation groups. It is implemented in the GAP computer algebra system. However, its outputs "are usually exponentially long"~\cite{fiat1989planning}, and that forbids practical computations for large groups (typically the sizes beyond $10^{30}$--$10^{40}$ are out of reach -- examples and analysis are given below).

{\bf Non-optimal pathfinding: "small support" Kaggle Santa 2023 methods (generality issue).} The methods which can deal with extremely large groups (e.g., $>10^{1000}$ in size like $33\times33\times33$ Rubik's cube) were demonstrated by top participants of the recent Kaggle challenge \href{https://www.kaggle.com/competitions/santa-2023}{Santa 2023}. However, the generality of that approach is unclear.
That is in contrast to AI methodology, which is, from the beginning general not even restricted to Cayley graphs. The idea of "Santa methods" is quite noteworthy: one should first find "small support elements" achievable from original generators and take them as new generators, then one simply runs beam search with Hamming distance heuristics to find path. This method becomes effective since for small support generators, Hamming distance, and true graph distance is quite related for obvious reasons. "Small support" is those permutations that change only a small number of positions. They are well known for Rubik's cube solvers -- "commutator" or "Pif-Paf" moves~\cite{mulholland2016permutation}, \href{https://math.stackexchange.com/q/4962862/21498}{M.Brandenburg post}. In mathematics, small support elements play an important role in deep works by Babai, Seress, Helfgott, et.al.~\cite{babai1988diameter, babai2004diameter, bamberg2014bounds, helfgott2014diameter, helfgott2019growth} (slides: \href{https://www.math.auckland.ac.nz/~conder/SODO-2012/Seress-SODO2012.pdf}{A.Seress},\href{https://simons.berkeley.edu/sites/default/files/docs/6206/symtalk.pdf}{H.Helfgott}). However, the question whether one can effectively find small support elements from given generators remains open, thus making the generality of "small support methods" unclear.

{\bf Non-optimal pathfinding: algorithms for specific families of generators.} 
Some algorithmic solvers can effectively find non-optimal solutions for some particular generator choices. It seems unlikely that one can produce similar algorithms for any generators. The prototypical example is the bubble sort algorithm, which solves the problem for the neighbor transposition generators of $S_n$. Algorithmic solvers exist for Rubik's cube of arbitrary sizes and other puzzle-related groups. The other families of generators  and algorithms are important in bioinformatics for estimating the evolutionary distance. These generators are related to flips of subsequences (see, e.g.,~\cite{Pevzner1995human2mice, Pevzner1999cabbage2turnip, bulteau2019parameterized}). A related example is \href{https://en.wikipedia.org/wiki/Pancake_sorting}{pancake sorting} (or prefix sorting) problem with the first algorithm proposed by Bill Gates et.al. \cite{gates1979bounds}.

{\bf AI-methods for pathfinding.} 
To the best of our knowledge, there is no systematic effort for Cayley graph pathfinding with AI methods except for our CayleyPy project. However, for some particular cases, for example, Rubik's cube, there are several approaches, e.g., based on genetic algorithms:~\cite{swita2023solving}.
The most notable are: DeepCube~\cite{mcaleer2019solving, agostinelli2019solving, khandelwal2024towards,agostinelli2024q} --- the first AI based successful solver for $3\times 3\times 3$ cube (\href{https://deepcube.igb.uci.edu/}{website}), the second one: EfficientCube \cite{takano2023selfsupervision}, some other:~\cite{brunetto2017deep,johnson2021solving,amrutha2022deep,noever2022puzzle,chasmai2022cubetr,bedaywi2023solving,pan2021fourier} which not achieve complete solution. Noteworthy idea~\cite{pan2021fourier} is to combine neural networks with the representation theory of the symmetric group ---  neural net  predicts the coefficients of the non-abelian Fourier transform for the distance function. The rationale is: observed sparsity (bandlimitedness) of Fourier transform of the common distance functions on $S_n$ ~\cite{swan2017harmonic}. The recent seminal achievement from S.Gukov's team: ~\cite{shehper2024makes} creates an effective AI-based pathfinding approach for an infinite group of Andrews-Curtis moves and resolves the  Akbulut-Kirby conjecture remained open for 39 years. AI methods are also used for pathfinding in the context of planning movements in obstacle-rich environments and road networks~ \cite{pandy2022learning, kirilenko2023transpath}.

After the first version of the present work, several new contributions appeared:   \cite{douglas2025diffusion}  proposed a novel diffussion-like approach to that problem, considered several classes of permutation and matrix groups and also demonstrated that the deep learning approach works successfully not only on Rubik's cube, but on general Cayley graphs up to certain sizes.
\cite{ZiarkoBortkiewiczZawalskiEysenbachMilos2025CRTR} provided solution of 333 Rubik's cube based on 
approach of latent representations. 

{\bf Cayley graphs: applications, diameters, random walks, open conjectures.}
Let us provide more general context on Cayley graph research.
Cayley graphs are fundamental in group theory~\cite{gromov1993geometric},\cite{tao2015expansion}, and have various applications: bioinformatics
~\cite{Pevzner1995human2mice, Pevzner1999cabbage2turnip,  bulteau2019parameterized}; processors interconnection networks~\cite{akers1989group, cooperman1991applications,heydemann1997cayley}; in coding theory  and cryptography~\cite{hoory2006expander,zemor1994hash,petit2013rubik}; in quantum computing~\cite{ruiz2024quantum,sarkar2024quantum,dinur2023good, acevedo2006exploring,gromada2022some}, etc.

Pure mathematical research on finite Cayley graphs focuses on diameter estimation and the behavior of random walks. In the case of the symmetric group $S_n$, two open conjectures are easy to formulate, but somewhat representative of the field: 

\begin{itemize}
    \item 
    {\bf Babai-like conjecture:} for any choices of generators diameter of  $S_n$ is $O(n^{2})$ (see, e.g., ~\cite{helfgott2014diameter},\cite{helfgott2019growth},\cite{helfgott2015random}); 

    \item 
    {\bf Diaconis ~\cite{diaconis2013some} conjecture:}  the mixing time for random walks is $O(n^{3}\log n)$ (again for any choices of generators).
\end{itemize}

The first one can be thought of as a particular case of the Babai conjecture~\cite{babai1992diameter}, \cite{tao2015expansion}, which predicts that diameter of simple groups is not as large as one may expect: $O(\log^c|G|)$ for some absolute $c$ --- in contrast to abelian groups where it can be $O(|G|)$. This conjecture represents an intriguing connection between an algebraic property of a group — simplicity, which is independent of the choice of generators — and a property of its Cayley graphs — having small diameter, which does depend on the generators.




Estimating diameters of Cayley graphs is a hard problem. One of its applications is latency estimation for the processor interconnection problem~\cite{akers1989group},\cite{cooperman1991applications}, \cite{heydemann1997cayley} since diameter corresponds to maximal latency. 
It is also a question of interest for community of puzzles enthusiasts --- like Rubik's cube. In the context of puzzles the diameter is often called the "God's number" --- the number of moves the best algorithm can solve the worst configuration. It is widely unknown for most of the puzzles. Its determination for the Rubik's cube $3\times 3 \times3$  for the standard choices of generators required significant efforts and computational resources~\cite{Rokicki2014Diameter}.  It is not known precisely for higher cubes, neither for $3\times 3 \times3$ Rubik's with \href{https://www.speedsolving.com/wiki/index.php?title=Metric#STM}{STM, ATM} metrics.
Precise conjectures on $4\times 4 \times4$  and $5\times 5 \times5$ Rubik's cube diameters appeared recently~\cite{hirata2024probabilistic}, collections of conjectures and estimates for many puzzles can be found on \href{https://www.speedsolving.com/wiki/index.php?title=God%27s_Algorithm#Table_of_God.27s_Numbers}{speedsolving website}. 

As we will discuss below, novel AI methods might provide insights into such questions through computational experiments.

\section{Reinforcement learning (RL) and graphs pathfinding}
The last subsection (\ref{sec-RLDD}) presents our new proposal for combining reinforcement learning (RL) training with pretraining based on the diffusion distance approach introduced in our previous work \cite{chervov2025machinelearningapproachbeats}, which is briefly reviewed in Section \ref{sec-DD}. Both methods rely on beam search, recalled in Section \ref{sec-Beam}. The first subsection (\ref{Bellman_section}) provides a brief overview of reinforcement learning and explains how it naturally fits into the graph pathfinding framework — a simple yet not thoroughly discussed 
perspective in the existing literature, to the best of our knowledge.



\subsection{RL setup for pathfinding and Bellman equation}
\label{Bellman_section}

Reinforcement learning is an important subfield
of AI, which is different from supervised and unsupervised learning in the following respects: it aims to optimize {\bf cumulative rewards}, and it requires the agent to, in a sense, "{\bf create its own training set}". The agent should choose the best strategies to "explore the space" and select training examples by himself, in contrast to supervised learning, where the training set is given a priori.
Here we will briefly discuss it from the graph theory perspective.

{\bf Bellman equation and shortest paths}. Let $d(s)$ be the \textit{distance function} — the length of the shortest path from the state $s$ to the target node $e$. Denote the set of all neighbors of $s$ as $N(s)$. 
It is obvious that, the distance function $d(s)$ satisfies the \textit{Bellman equation}: 

\begin{equation}
\label{bel_eq}
d(s) = 1 + \min\limits_{t \in N(s)} d(t)  
\end{equation}
and the \textit{boundary condition} 

\begin{equation}
\label{boundary_cond}
d(e) = 0.
\end{equation}

The equation is elementary but plays an important role in reinforcement learning and its deep learning generalizations discussed below.

The Bellman equation implies a simple proposition, which is easy to see inductively considering neighbor nodes to $e$, then neighbors of neighbors, etc. (actually, this is how the breadth-first search algorithm finds the shortest path).

\begin{proposition}
    For any finite connected graph $G$, the only solution to the equations~\eqref{bel_eq} and~\eqref{boundary_cond} is given by the distance function $d(s)$. Moreover, the only solutions to Bellman's equation~\eqref{bel_eq} (without boundary condition) correspond to the distance to some sets of nodes in $G$ (up to adding a constant). That set can be recovered as minimums of $d(s)$. 
\end{proposition}

{\bf RL "equivalence" to graph pathfinding.} 
Here we remind the standard form of Bellman equation for the \textit{value function} $v_*(s)$  (\cite[p.62]{sutton2018})  and relate it to the form given above.  Taking any action $a$ in state $s$ the agent receives a reward (penalty) $r(s, a)$,  and denote \textit{ discount factor} to be  $\gamma \in(0,1]$, then   \textit{Bellman optimality equation}:


$$
 v_*(s) = \max\limits_a \big\{r(s, a) + \gamma \mathbb E_{s'} v_*(s')\big\}
 $$
\noindent
Considering environment to be deterministic,  $\mathbb E_{s'} v_*(s') = v_*(a\circ s)$ and taking $r(s, a)  = -1$, one gets:
\begin{equation}
 \label{eq:v-star-bellman}
 v_*(s) =
 -1+ \gamma\max\limits_a \big\{v_*(a\circ s)\big\}.
 \end{equation}
\noindent
Now observe that~\eqref{bel_eq} turns into~\eqref{eq:v-star-bellman} if $\gamma = 1$ and $d(s) = -v_*(s)$. By definition of the value function $v_*(e) = 0$ which is the same as~\eqref{boundary_cond}.

{\bf Solving Bellman - simple iteration method (dynamic programming).} 
Bellman equation on unknown function $d$ has the form $d = F(d)$, thus to solve it one can use method of simple iteration:
$d_{k+1} = F (d_{k} )$. 
Thus, solving Bellman equation is conceptually rather simple - 
starting with some initialization $d_0(s)$, the new approximation for the distance function is iteratively updated by
by the right hand side of the Bellman equation (method is also called dynamic programming in RL context), i.e.:
$$
d_k(s) = 1+\min_{t \in N(s)} d_{k-1}(t), \quad d_{k}(e) =0.
$$
\begin{algorithm}[H]
 \caption{Simple iteration (Dynamic programming)}
 \begin{algorithmic}[1]
 \STATE \textbf{Hyperparameters:} $\epsilon > 0$ (tolerance), $\alpha \in (0, 1]$ (learning rate), $N$ (maximum iterations)
 \STATE Initialize $d_0(s)$ for all states $s \in \mathcal S$ (e.g., by zeros)
 \FOR{$i = 0$ to $N$}
 \STATE $d_{i+1}(e) \leftarrow 0$
 \FOR{each state $s \ne e$}
 \STATE 
 $d_{i+1}(s) \leftarrow  \alpha(1 + \min\limits_a d_i(a\circ s)) + (1-\alpha)d_i(s)$
 \ENDFOR
 \STATE
 $c_i \leftarrow \mathrm{corr}(\boldsymbol d_i, \boldsymbol d)$
 \IF {$\Vert \boldsymbol d_i - \boldsymbol d\Vert  < \epsilon$}
 \STATE \textbf{break}
 \ENDIF
 \ENDFOR
 \STATE \textbf{Output:} Number of iterations $i$, vector of correlations $\boldsymbol c$ 
 \end{algorithmic}
 \label{alg:DP}
 \end{algorithm}

{\bf Examples of solving Bellman equation. Initialization dependence.}
Let us illustrate solving Bellman equation on Cayley graphs and demonstrate some unexpected dependence on initializations.
We observe that if initialization is positive convergence takes not more than diameter steps (it is no longer true for initialization with negative values). However convergence might not be monotonic - depending on initialization, it might happen that first
solution deviates from the final result, and only after several steps, begins to converge again. 

For small graphs such as \href{https://www.kaggle.com/code/fedmug/bellman-rc2}{Pocket Cube} or \href{https://www.kaggle.com/code/fedmug/rl-dynamic-programming-on-small-lrx}{LRX with $n\leqslant 10$} the true distances $d(s)$ can be easily calculated via BFS for each state. Therefore, we can use these true distances as stopping criteria: when $\|d_n(s) - d(s)\| < \varepsilon$, we are finished. In addition, we measure performance on each iteration by calculating the correlations between $d(s)$ and $d_k(s)$.

The main question of interest here was how the initialization influences the performance. That is important to understand in view of our proposal to use initialization coming from the diffusion distance.
We tried different initializations $d_0(s)$:

\begin{itemize}
 \item 
 zero initialization $d_0(s) = 0$ for all $s$
 \item 
 initialize each of $m$ layers by a specific value (e.g., first $k$ layers by true distances, last $m - k$ layers by zeros, $1 \leqslant k < m$)
 \item
 Manhattan distance to the target: 
 $d_0(s) = \sum \vert s_k - k \vert$
 \item 
 Hamming distance to the target:
 $d_0(s) = \vert \{k\colon s_k \ne k\}\vert$

\item 
 random integers form $0$ to $m$ or random values from $\mathcal N(0, \sigma^2)$

\item 
outputs of a model predicting diffusion distances (see next subsection or formula~\eqref{eq:dd})
 \end{itemize}

\begin{figure}[ht]
    \centering
    \includegraphics[width=1.0\linewidth]{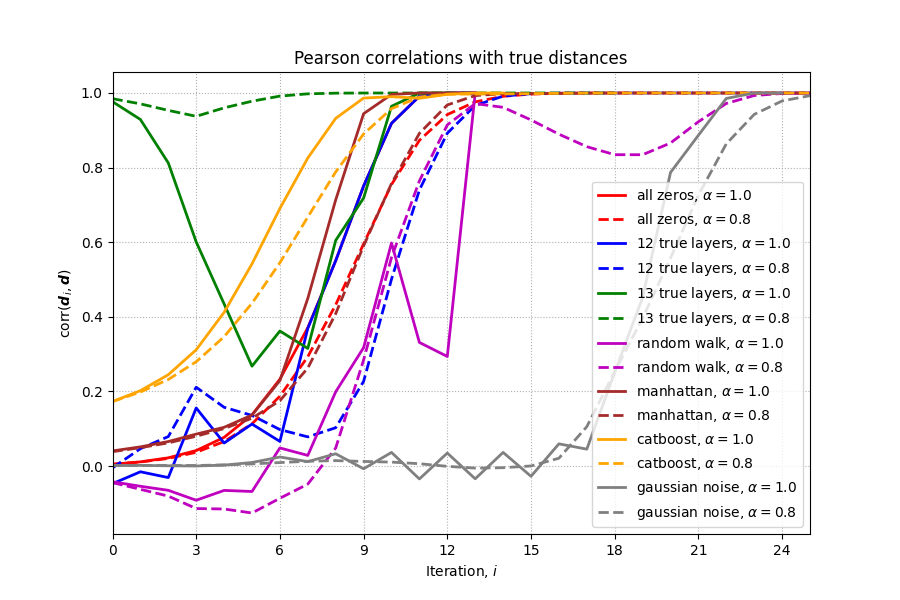}
    \caption{Pearson correlation between $d_i(s)$ and $d(s)$ on each iteration of the DP algorithm, $\epsilon = 10^{-3}$ (Pocket Cube)}\label{fig:pearson}
\end{figure}

These experiments revealed interesting phenomena presented in figure \ref{fig:pearson}. 
For some specific initialization the performance first degrades, and only after that achieves the convergence to the true distance which is guaranteed by the general theory. The number of steps for convergence does not exceed the diameter of the graph if initialization is positive, however, it can be larger in presence of negative values. Initialization coming from the machine learning model ("catboost") trained on diffusion distance (same as "warm-up step" in our proposal) improves the convergence, though not in a radical manner.

{\bf Deep Q-learning algorithm.} Tabular methods like DP or Q-learning are not applicable in situations where the graph is too large. A family of deep Q-learning methods like~\cite{mnih2013playing} emerged to resolve this issue. Typically, these algorithms leverage a neural network to predict targets like right-hand sides of~\ref{eq:v-star-bellman} or~\ref{bel_eq}:

$$
    f_{\boldsymbol \theta}(s) = r + \max_a Q_{\boldsymbol \theta}(s, a).
$$

The algorithm~\ref{alg:deepQ} is a deep Q-learning method designed to solve the Bellman equation~\ref{bel_eq} (here we assume that $s$ is an embedding of a graph node suitable for input of the neural network $f_{\boldsymbol \theta}(s)$). Typically, the choice of the training subset of nodes is made by walking over the graph, combining both random steps and steps guided by the model. After the training is completed, the output of the model $f_{\boldsymbol \theta}(s)$ should hopefully serve as a good approximation to the true distance from $s$ to $e$.

\begin{algorithm}[t]
 \caption{Deep Q-learning}
 \label{alg:deepQ}
 \begin{algorithmic}[1]
 \STATE \textbf{Hyperparameters:} $N$ (number of epochs), $M$ (subset size)
 \STATE Initialize weights $\boldsymbol \theta$ of the neural network $f_{\boldsymbol \theta}(s)$
 \FOR{$i = 0$ to $N$}
 \STATE select a random subset of nodes $S_i$, $|S_i| = M$, $e\notin S_i$
 \FOR{each state $s \in S_i$}
 \STATE 
 $d_i(s) \leftarrow  1 + \min\limits_{t \in N(s)}f_{\boldsymbol \theta}(t)$
 \ENDFOR
 \STATE calculate MSE loss $\mathcal L(\boldsymbol \theta) = \frac 1M \sum\limits_{s\in S_i}\big(d_i(s) - f_{\boldsymbol \theta}(s)\big)^2$
 \STATE calculate $\nabla_{\boldsymbol \theta} \mathcal L(\boldsymbol \theta)$ and do the gradient step
 \ENDFOR
 \end{algorithmic}

 \end{algorithm}

{\bf Graph path finding.}
To find a path from a given node $s$ to the target  $e$, one needs to run a greedy search algorithm, which, starting from node $s$, inspects all its neighbors and evaluates distance function for them, then chooses the node with the minimum value of the distance function and makes a move to that node. The process is repeated until the number of the destination node found at the limit of the step is exceeded. If the distance function is the true distance on the graph (the shortest path length), such an algorithm finds the shortest path optimally (least possible operations). However, the predictions of the neural network $f_{\boldsymbol \theta}(s)$ from the deep Q-learning algorithm~\ref{alg:deepQ}, are typically not very precise, and the greedy algorithm will be stuck in local minima. To overcome that difficulty, one used more advanced graph search algorithms like beam search, A-star, etc. 

A guided beam search approach is described in algorithm~\ref{alg:beam_search}; on each iteration it maintains $W$ states with lowest estimations to the target state. If algorithm~\ref{alg:beam_search} can find the path to the target in $N$ iterations, it returns the length of this path.

\subsection{Diffusion distance pathfinding (CayleyPy-1)}\label{sec-DD}
Let us briefly remind the alternative pathfinding approach proposed in our previous paper \cite{chervov2025machinelearningapproachbeats}. Its efficiency has been demonstrated on the Rubik's cube, where significantly outperforms all possible competitors, providing shorter solutions than any existing approach for $3\times 3\times3$, $4\times 4\times 4$, $5\times 5\times5$ cubes. In some sense, it is simpler and more computationally efficient.

\begin{algorithm}[t]
 \caption{Guided beam search pathfinding}
 \label{alg:beam_search}
 \begin{algorithmic}[1]
 \STATE \textbf{Hyperparameters:} $s_0$ (starting state), $M$ (step limit), $f(s)$ (distance estimator), $N(s)$ (neighbors of $s$), $W$ (beam width)
 \STATE $B \leftarrow \{s_0\}$
 \FOR{$i = 0$ to $M$}
    \STATE $B' \leftarrow \varnothing$
    \FOR{each state $s \in B$}
        \IF{$s = e$}
         \STATE \textbf{return} $i$
        \ENDIF
        \STATE $B' \leftarrow B' \cup N(s)$
    \ENDFOR
    \STATE $B' \leftarrow B' \setminus B$
    \STATE sort states $s \in B'$ by the distance estimations: 
    $$
    f(s_{(1)}) \le f(s_{(2)}) \le \ldots \le f(s_{(|B'|)})
    $$ 
    \vspace{-5mm}
    \STATE $B\leftarrow \{s_{(1)}, \ldots, s_{(\min\{W, |B'|\})}\}$
 \ENDFOR
 \end{algorithmic}
 \end{algorithm}

The idea is to work not with the true distance on the graph but with the diffusion distance, which is the length of the random path. The rigorous definition of the diffusion distance is given by the formula~\eqref{eq:dd} in the supplementary section.




The diffusion distance is easy to estimate by the number of steps of random walks. Generating random walks is much faster than computing neural network predictions. Thus, the training data generation is much more efficient than in Q-learning approaches, though the training data might be of poorer quality. 



The core of our approach consists of three steps:

\begin{enumerate}
    \item {\bf Generation of the training set via random walks.} 
    Generate $N$ random walk trajectories starting from the selected node $e$. Each random walk trajectory consists of up to $K_{\text{max}}$ steps, where $N$ and $K_{\text{max}}$ are integer parameters of the method. If a random walk visits node $s$ for the first time at step $k$, we store the pair $(s,k)$ in our training set. 
    
    \item {\bf Training a model.}  
    The generated set of pairs $(s,k)$ serves as the training data for a supervised machine learning model $f(s)$, typically a neural network.
    For a given node $s$ this model predicts the estimated diffusion distance from $s$ to the target $e$.

    \item {\bf Guided beam search.} Here we apply the algorithm~\ref{alg:beam_search} to find a path from a given state $s$ to the target node $e$. The model $f(s)$ trained on the previous step provides heuristics on where to make the next steps, while the beam search technique compensates for any possible incorrectness in the predictions of the neural network. Having fixed a positive integer $W$ ("beam width" or "beam size"), on each step we select $W$ nodes that are closest to the destination according to the neural network. Then we check $f(s)$ for all neighbors of these best $W$ states, drop duplicates, and repeat the process until the destination node is found or the step limit is exceeded.
  
\end{enumerate}

The last step is common for any method, whether it is based on Q-learning or diffusion distance. One uses a neural network predictions as a heuristic function and graph search algorithms to find a path. We are strongly in favor of beam search since it is quite simple, but has proven to be the most effective for us, while the DeepCube team~\cite{mcaleer2019solving, agostinelli2019solving, khandelwal2024towards,agostinelli2024q} uses the A-star algorithm and develops its modifications.

A simple but useful modification of the diffusion distance method leverages \textbf{non-backtracking} which is applied during steps~(1) and~(3). On both occasions, we store the history states visited in the $m$ previous steps to avoid returning to them. The history of non-backtracking random walks contains $O(mN)$ states, and all trajectories are not allowed to visit any of these states and their neighbors in the next step. For non-backtracking beam search we do not include the states from the history on line 9 of the Algorithm~\ref{alg:beam_search}.

\subsection{Beam search effectivity. Toy model of research community }\label{sec-Beam}

To some extent, the success of the proposed above is due to beam search and our efficient implementation supporting large beams. We started with a modified greedy search, then moved to Metropolis and A-star, before finally finding relief in beam search. The larger the beam size, the more effective beam search becomes (probability to find path increases and paths become shorter). To maximize beam sizes, we developed an original and efficient PyTorch implementation capable of supporting beam sizes with millions of nodes and beyond (\href{https://www.kaggle.com/code/alexandervc/beamsearch-basic}{basic}, \href{https://github.com/iKolt/cayleypy/blob/main/cayleypy/cayley_graph.py#L447}{advanced} versions). On one hand, beam search is arguably one of the simplest methods. On the other hand, there are several key aspects worth highlighting.

The primary reason we need beam search is to avoid the problem of local minima. Greedy search can get stuck in them, whereas beam search can bypass them entirely if the beam size is larger than the depth of the local minima. It is similar to how an elephant wouldn't even notice small dips in the terrain, while for an ants, they could be an insurmountable obstacles.

Another important aspect of beam search is that increasing the beam size increases both computation and memory consumption linearly. Typically, memory -- not computation time -- is the main bottleneck. However, maximizing the beam size is crucial for performance.
Moreover, there is no simple way to reproduce the results of beam search with a beam size of $B+1$ using any algorithm that only consumes memory equivalent to beam size $B$.

To explain this point, we will use the analogy with the work of the research community.
Imagine two researchers working together on a creative problem (a beam size of 2). The key point is the exchange of ideas: if one researcher makes a breakthrough, they share it with the other, allowing both to continue their search from the breakthrough point. Without this exchange, the second researcher might get stuck in a dead end. This sharing of ideas helps avoid wasting time in unproductive directions.
The mechanism described above explains why increasing the beam width is so effective: it acts as an exchange of ideas, focusing on breakthroughs rather than following unproductive paths.
Essentially, this is what beam search does: it selects the top nodes across the entire beam neighborhood rather than picking the best few for each individual node. This means that if the neighbors of one node are significantly closer to the destination (analogous to a breakthrough), those nodes are retained in the beam, while the neighbors of less successful nodes are discarded. In other words, the descendants of some nodes do not survive into the next generation. A kind of natural selection process.


These observations show that community size has a disproportionately large impact on research productivity. Halving the number of researchers does not simply halve the output — it causes a super-linear (often exponential) decline. In contrast, for routine tasks, a 50\% reduction in resources usually yields about a 50\% reduction in output. The same principle applies to beam search: the beam size is crucial, and there is no simple way to achieve the performance of a beam size $B$ using a smaller beam.

This logic also explains why evolutionary or nature-inspired global optimization methods can sometimes be successful. They rely on the same fundamental mechanism as beam search but offer better control over the diversity within the beam.
In some cases, it is important not to immediately jump to breakthroughs made by others but to continue exploring independent directions, as this might lead to even greater discoveries. For example, if medical doctors would stop doing their job to start follow breakthroughs in mathematics, it would lead to disastrous results — diversity is crucial in the long run.
The balance between "everyone follows the latest breakthrough" (as in beam search) and maintaining diversity is delicate. There is no universal solution, as different tasks require different trade-offs. Exploring various methods on Cayley graph pathfinding tasks is an interesting direction.

\subsection{Combining diffusion distance with deep Q-learning} \label{sec-RLDD}
The diffusion distance approach described above is quite simple and efficient; however, there are practical and theoretical disadvantages. The practical side consists of somewhat surprising phenomena that enlarging the training set (generating more random walks) does not always improve the quality of the model; such stagnation was observed for Rubik's cube cases in our previous paper, and it is even more prominent for LRX graphs. It is actually related to the quite clear theoretical fact: diffusion distance is not monotonic function of the true distance.
This problem is not only for random walks estimation, but also for the diffusion distance itself (e.g., see computed theoretically \href{https://www.kaggle.com/code/fedmug/lrx-diffusion-distance-analysis}{analysis}).
So, diffusion distance is easy to estimate, but it is not precise. On the other hand, DQN and Bellman's equation allow us to approximate the true distance, but for the price of more heavy computations. 
The idea of the proposal below merges lightweight diffusion distance with a heavier and more precise Bellman equation approach, and at avoiding the sparse reward problem.

Our proposal looks as follows.

{\bf Part 1. Warm-up diffusion distance training.} Apply diffusion distance approach for pre-training the neural network, i.e. steps (1) and (2) above: generate training data via random walks, and train a neural network $f_{\boldsymbol \theta}(s)$ to predict the number of steps of the random walk.

{\bf Part 2. Modified DQN training.} We use the deep Q-learning algorithm~\ref{alg:deepQ} with two enhancements:

\begin{itemize}
    \item 
    the random subset of nodes consist of all states visited by $N$ random walks of size $K_{\max}$;
    \item 
    the result of the Bellman update $1 + \min\limits_{t \in N(s)}f_{\boldsymbol \theta}(t)$ is clipped to the interval $[0, k]$ where $k$ is the time step when the state $s$ was visited for the first time.
\end{itemize}

The pseudo code for this modified version of DQN is presented in Algorithm~\ref{alg:deepQ_modified}.






{\bf Part 3. Guided beam search.} Here we apply step (3) from the diffusion distance approach described above (Algorithm~\ref{alg:beam_search}).


\begin{algorithm}[t]
 \caption{Modified deep Q-learning}
 \label{alg:deepQ_modified}
 \begin{algorithmic}[1]
 \STATE \textbf{Hyperparameters:} $M$ (number of epochs), $N$ (number of random walks), $K_{\max}$ (length of random walks), $f_{\boldsymbol \theta}(s)$ (distance estimator)
 \FOR{$i = 0$ to $M$}
 \STATE Generate $N$ random walks of size $K_{\max}$ from the target node $e$
 \STATE Collect dataset $\mathcal D = \{(s, k)\}$ of states $s$ visited by random walks at time $k$
 \FOR{each $(s, k) \in \mathcal D$}
  \STATE 
  $d(s) \leftarrow  1 + \min\limits_{t \in N(s)}f_{\boldsymbol \theta}(t)$ (update based on Bellman equation~\eqref{bel_eq})
  \STATE $d(s) \leftarrow \min\{k, \max\{0, d(s)\}\}$ (clipping)
 \ENDFOR
 \STATE Calculate MSE loss $\mathcal L(\boldsymbol \theta) = \frac 1{|\mathcal D|} \sum\limits_{\mathcal D}\big(d(s) - f_{\boldsymbol \theta}(s)\big)^2$
 \STATE Calculate $\nabla_{\boldsymbol \theta} \mathcal L(\boldsymbol \theta)$ and do the gradient step
 \ENDFOR
 \end{algorithmic}

 \end{algorithm}

There are several advantages of the proposed method. First, it avoids the sparse reward problem: since we always start from the selected state, we can reliably clip overestimated predictions. Second, the warm-up phase helps avoid the long initialization typical for standard DQN, where targets from the Bellman equation are noisy because the network starts with random weights. In our approach, the targets are meaningful from the beginning: in the first phase, they come from random walk steps; in the second phase, the network itself produces useful targets, as it has already been pre-trained. Finally, we use non-backtracking random walks for a clear reason: the number of steps in such walks better reflects the true distance. Conceptually, this relates to the well-known phenomenon "Non-backtracking random walks mix faster"~\cite{alon2007non}.

\section{Analysis and results}

DeepMind's success in mastering the game of Go, which has approximately $10^{170}$ possible states, poses an inspiring challenge: achieving similar breakthroughs in other domains, such as Cayley graphs. At present, we are still a distance from this goal. However, many promising ideas have yet to be implemented in our CayleyPy project, and with continued development, reaching such a milestone may be possible in the near future.

In this study, we demonstrate successful pathfinding in very large LRX Cayley graphs of $S_n$. We use its conjecturally longest element~\eqref{eq:longest} of expected length $n(n-1)/2$ as a benchmark.  Our zero prior knowledge approach can find the paths from $\ell_n$ to $e$ for $n \leqslant 44$. This is a significant improvement over the performance of the GAP computer algebra system, which can only handle up to around $n=20$, and with much longer path lengths and runtimes.  Introducing some domain knowledge (see Section 3.2 below) drastically improves the performance to $n=100$ or even more.


\subsection{"Overcome the GAP" with zero prior knowledge AI }
Here we demonstrate that the diffusion distance method significantly outperform classical methods of computer algebra system GAP. 

\begin{table}[ht]
    \centering
    \begin{tabular}{|c|c|c|c|c|c|}
        \hline
        $n$  &  GAP length   & $\frac{n(n-1)}2$            &   DD length    \\
        \hline
        9    &  41     & 36               &  36   \\
        \hline
        10   &    51   & 45               &  45       \\
        \hline
        11   &     65      & 55             &  55     \\
        \hline
        12   &    78       &  66          & 66     \\
        \hline        
        13   &    99       &  78           &   78       \\
        \hline        
        14   &   111       &  91           &   91     \\
        \hline        
        15     &   268      &  105         &  105    \\
        \hline        
        16     &    2454    &  120         &  120     \\
        \hline        
        17   &     380     &  136         & 136    \\
        \hline        
        18   &     20441   &   153        & 153    \\
        \hline        
        19   &      3187      & 171        & 171    \\
        \hline      
        20   &     217944    & 190        & 190      \\
        \hline 
        21   &   -           & 210     & 210        \\
        \hline           
    \end{tabular}
    \captionsetup{skip=10pt} 
    \caption{Comparison of GAP and diffusion distance (DD) method for the conjecturally longest element of LRX Cayley graph with expected length $n(n-1)/2$.}
    \label{tab:example}
\end{table}

For $n=20$, the  GAP timing is 41min 18s,  while our methods can find results much faster.
For example, the diffusion distance method with simple neural network requires only 3m 48s of calculations on GPU P100 (\href{https://www.kaggle.com/code/alexandervc/lrx-cayleypy-rl-mdqn?scriptVersionId=224270083}{notebook version 423}). GAP benchmarks have been performed \href{https://www.kaggle.com/code/avm888/group-elements-decomposition-gap-limits}{here}.

\begin{table}[ht]
    \centering
    \begin{tabular}{|c|c|c||c|c|c|}
\hline
$n$  & $\frac{n(n-1)}{2}$ & DD length &
$n$  & $\frac{n(n-1)}{2}$ & DD length \\
\hline\hline
22 & 231 & 231  & 31 & 465 & 478 \\
\hline
23 & 253 & 253  & 32 & 496 & 508 \\
\hline
24 & 276 & 276  & 33 & 528 & 642 \\
\hline
25 & 300 & 300  & 34 & 561 & 667 \\
\hline
26 & 325 & 325  & 36 & 630 & 834 \\
\hline
27 & 351 & 351  & 38 & 703 & 759 \\
\hline
28 & 378 & 378  & 40 & 780 & 926 \\
\hline
29 & 406 & 408  & 42 & 861 & 909 \\
\hline
30 & 435 & 437  & 44 & 946 & 1060 \\
\hline
\end{tabular}
    \captionsetup{skip=10pt} 
    \caption{Lengths of solutions produced by diffusion distance (DD) method for the conjecturally longest element of LRX Cayley graph with expected length $n(n-1)/2$.  }
    \label{tab:example_nn}
\end{table}

The \texttt{DD length} columns in Tables~\ref{tab:example} and~\ref{tab:example_nn} are aggregated results from experiments with different guiding models based on gradient boostings and neural networks.
All of them solve cases with $n\le 28$ quite stably; moreover, after some hyperparameter tuning, they find the shortest paths. Going beyond the limit $n=28$ requires some effort. For $n\le 44$ a path can be occasionally found by multiple relaunches of the pipeline, however, its length is no longer optimal.

\subsection{X-trick} Surprisingly, adding one line of code allows us to drastically increase the size of solvable graph, from $S_{40}$ to $S_{100}$. Namely, add the following condition in line~9 of the beam search algorithm~\ref{alg:beam_search}: if $s_0 < s_1$ then do not include $X\circ s$ in $N(s)$.
The logic is straightforward: if the first two elements of the permutation are already sorted, it makes little sense to swap them by applying the generator $X = (0 \ 1)$. 

This X-trick turns out to be a game changer. By adding it to the pipelines that succeeded for $n$ about $40$, we were able to find paths for $n=100+$ (and potentially even larger values, as we were limited by the 16GB RAM limit). This represents a shift from working with Cayley graphs of size $10^{30+}$ to $10^{150+}$.  This trick worked universally across model types, including perceptrons, gradient boosting machines, and convolutional neural networks. The key challenge now is how to eliminate such prior knowledge and generalize the method to other types of generators. However, X-trick may prevent the discovery of optimal paths, especially for small values of $n \le 15 $.

Examples of lengths found by the diffusion distance method with X-trick are presented
in Table~\ref{tab:lens_upto100} below:

\begin{table}[ht]
    \centering
    \begin{tabular}{|c|c|c|c|c|c|}
        \hline
        $n$   & 100 & 90   & 80  & 70 & 60  \\
        \hline
        Solution length        & 16632 & 33459  & 7510  & 8115 & 3650 \\
        \hline
        Diameter (proposed) &  4950   & 4005  & 3160  & 2415 & 1770 \\
        \hline
        Time         &  8h 43m & 3h 57m   & 1h 5m  & 52m & 3h  \\
        \hline
        Notebook version   &   255   &  242     & 241    & 233  & - \\
        \hline
    \end{tabular}
    \captionsetup{skip=10pt} 
    \caption{Lengths of solutions for different $n$ obtained by ML model followed by beam search with X-trick. (Not completely zero prior knowledge approach). }
    \label{tab:lens_upto100}
\end{table}

Computation for $n=100$ took 8h 43m 53s GPU P100. Beam width $ {2}^{20}$, 30 epochs of training ( \href{https://www.kaggle.com/code/alexandervc/lrx-cayleypy-rl-mdqn?scriptVersionId=223954525}{Notebook version 255}). For other cases we used beam size $\le {2}^{18}$, that is why for $n=90$ length is much larger. The data for $n=60$ is obtained by the gradient boosting model, with slightly better lengths, but more timing, in the other cases single layer MLP model used. In all cases we decompose the conjecturally longest element of expected length $n(n-1)/2$. 

\subsection{Diffusion distance + modified DQN}
We observe that replacing the training stage with the DQN algorithm~\ref{alg:deepQ_modified} gives a consistent improvement over the diffusion distance approach; however, the effect is not very large. Below we provide two illustrative experiments.

\begin{table}[ht]
    \centering
    \begin{tabular}{|c|c|c|c|c|}
        \hline
        $n$  & epochs: $5+0$  & epochs: $50+0$   & epochs: $5+50$   & epochs: $50+50$ \\
        \hline\hline
        6 &  0.955 & 0.99  & 0.99  & 0.99 \\
        \hline
        8 &  0.954 & 0.969 & 0.98   & 0.982 \\
        \hline
        10 & 0.883  & 0.923 & 0.939   &  0.938 \\
        \hline
    \end{tabular}
    \captionsetup{skip=10pt} 
    \caption{Spearman correlation between the true distances and the predictions of the neural network trained for $A$ epochs during the warm-up diffusion distance phase and $B$ epochs during the DQN phase (epochs: $A+ B$). Additional training with DQN leads to improved results.}
    \label{tab:DQN_small}
\end{table}

The first experiment (see Table~\ref{tab:DQN_small}) focuses on relatively small groups, where the true graph distance can be computed by brute force and compared with the neural network's predictions. For small group sizes, the method learns the graph distances almost perfectly. However, for larger groups, the correlation between predictions and true distances becomes less accurate. Still, additional training using the modified DQN always improves upon the initial warm-up phase based on the diffusion distance procedure.  

\begin{table}[ht]
    \centering
    \resizebox{0.95\textwidth}{!}{
    \begin{tabular}{|c|c|c|c|c|c|}
        \hline
        Warm-up epochs  & DQN epochs  & success rate & shortest path & median path \\
        \hline\hline
        30 & 200 & 80\%  &  382 & 402  \\
        \hline
        30 &  0 & 70\%  & 464 &  650  \\
        \hline
        100 &  0 & 70\%  & 436 & 516  \\
        \hline
        0 &  230 & 20\%  & 434 & 471  \\
        \hline
    \end{tabular}}
    \captionsetup{skip=10pt} 
    \caption{Path lengths for decomposing the longest element in the LRX Cayley graph of $S_{28}$.
    Results in each row are based on 10 independent runs. Additional training with DQN improves both the average path length and the success rate of finding a path. Pure DQN training alone performs poorly. }
    \label{tab:DQN_28}
\end{table}

The second experiment concerns the larger group $S_{28}$ with LRX generators. We train the model and use beam search to find a path, where the path length serves as the performance metric. The results provided in Table~\ref{tab:beam_size} confirm that additional training with DQN improves performance. The combination of $30$ warm-up and $200$ DQN epochs produces one of the best outcomes observed in this setting. The median (shortest) path length $402$ ($382$) is very close to the expected ideal length of $378$. For each run, models were retrained from scratch; thus, due to the randomness of the walks, the resulting models differ significantly. The analysis can be found in the Kaggle \href{https://www.kaggle.com/code/alexandervc/lrx-cayleypy-rl-mdqn}{notebook}. 

\subsection{Non-backtracking}
We exploit non-backtracking idea in two ways - first we can use non-backtra-cking random walks, and the second we can use non-backtracking in beam search. The second way shows quite a good effect. 



\begin{table}[t]
    \centering
    \begin{tabular}{|c|c|c|c|}
        \hline
        history size, $m$ & success rate & median length & min length \\
        \hline\hline
        2   & $40.41\%$ & $476$ & $380$ \\
\hline
4   & $40.10\%$ & $474$ & $382$ \\
\hline
8   & $40.32\%$ & $471$ & $386$ \\
\hline
16  & $37.40\%$ & $470$ & $382$ \\
\hline
32  & $37.47\%$ & $470$ & $378$ \\
\hline
40  & $36.16\%$ & $480$ & $390$ \\
\hline
48  & $37.39\%$ & $465$ & $380$ \\
\hline
128 & $36.45\%$ & $476$ & $382$ \\
\hline
254 & $37.85\%$ & $468$ & $382$ \\
\hline
400 & $37.40\%$ & $474$ & $384$ \\
\hline
512 & $36.58\%$ & $476$ & $384$ \\
\hline    
    \end{tabular}
    \captionsetup{skip=10pt} 
    \caption{Performance of the DD method with non-backtracking random walks ($n=28$). }
    \label{tab:nbt}
\end{table}

\begin{table}[ht]
    \centering
    \begin{tabular}{|c|c|c|c|c|c|}
        \hline
        Beam   &  Solution Length                  \\
       Width   &  Non-backtracking RW    \\
        \hline
        $2^{10}$    &       None           \\
        \hline
        $2^{12}$   &       252             \\
        \hline      
        $2^{14}$    &       202             \\
        \hline    
        $2^{16}$    &       194            \\
        \hline 
        $2^{18}$    &       194             \\
        \hline  
        $2^{20}$     &       190 (Ideal)    \\
        \hline        
    \end{tabular}
    \captionsetup{skip=10pt} 
    \caption{Influence of beam width and non-backtracking random walks for solving the longest element of the LRX Cayley graph $S_{20}$.     }
    \label{tab:beam_size}
\end{table}


The theoretical reason why non-backtracking is helpful is clear for the random walks. Imagine our graph is a tree, then  for a non-backtracking  random walk, the number of steps coincides with the true distance (length of the shortest path). Thus diffusion distance coincides with  the standard
distance. For complicated graphs, this is not the case, but still, it forces the diffusion distance to be closer to true distance and to have less variability. Here by variability we mean the situation when the same node can be achieved in a different number of steps. This is equivalent to having noise in the training data which makes models less precise. Similar for beam search non-backtracking mitigates stagnation
in local minimums. 

\begin{table}[ht!]
\centering
\begin{tabular}{|c|c|c|c|}
\hline
history size $m$ & success rate & median length & min length \\
\hline
2     & $17.65\%$ & $430$ & $384$ \\
\hline
4     & $17.08\%$ & $432$ & $390$ \\
\hline
8     & $21.37\%$ & $440$ & $382$ \\
\hline
16    & $30.07\%$ & $446$ & $388$ \\
\hline
32    & $38.55\%$ & $473$ & $384$ \\
\hline
64    & $43.29\%$ & $495$ & $380$ \\
\hline
128   & $56.95\%$ & $517$ & $384$ \\
\hline
256   & $60.61\%$ & $588$ & $384$ \\
\hline
512   & $67.33\%$ & $582$ & $384$ \\
\hline
1024  & $69.71\%$ & $663$ & $382$ \\
\hline
\end{tabular}
\caption{Performance of the DD method with non-backtracking beam search ($n=28$). The path is found more often as the history size grows. }
\label{tab:beam_ban}
\end{table}

Injecting non-backtracking into the beam search stage (Algorithm~\ref{alg:beam_search}) allows to increase the success rate: the greater is the number of banned states not included in the beam, the higher is the probability of finding a solution (see Table~\ref{tab:beam_ban}). On the other hand, the median solution length tends to increase when the history size is large. But that cannot be interpreted as performance degradation, rather it means that pipeline can solve more difficult cases, and hence getting large lengths.

\subsection{Architectures and parameter dependence}
We observe that rather simple models like MLP and gradient boostings currently perform better than more advanced ones: transformers, CNNs, etc. An exception is one of our transformer models that can find paths up to $n=100$, however, it is not a zero prior knowledge model since it uses features specific to LRX generators. 

The dependence on various parameters was explored and discussed in our previous paper ~\cite{chervov2025machinelearningapproachbeats} for the case of the Rubik's cube group. The beam size proved to be one of the most important parameters: the length of solution is almost linearly improving on logarithm of beam size. Another observation is that it is more important for a neural network to be deep than wide, however, the effect of depth is better seen for large groups. We also noted the performance stagnation as size of the training set increased.  

For the LRX Cayley graph these phenomenons are less strongly expressed, or even negligible. 
Most experiments with deep neural networks have not lead to significant improvements over simple one layer perceptron with 128 neurons. In particular, one can train models in just a single epoch within a minute, e.g., for $S_{20}$. The beam size effect is still present but not as pronounced (see table~\ref{tab:beam_size}).

{\bf CNN.}
Convolutional neural networks have demonstrated a slower training process. Their results are similar to those of perceptrons, e.g., length of the solution for $n=16$ is 122 (ideal is 120), for $n=20$ is 216 (ideal is 190). The basic version of the architecture uses two 1-d CNN layers with batch normalization and dropout to prevent overfitting. One of the successful variations is using temporal gated CNN. The first part of the architecture is the residual block - several convolutional blocks with gating based on sigmoid activation function. The second part of the architecture is stacking several layers of residual blocks with gradually increased dilation to increase the size of the receptive field.
Notebooks: \href{https://www.kaggle.com/code/artgor/encoding-as-permutation-matrix-1d-cnn}{1}, \href{https://www.kaggle.com/code/artgor/encoding-as-permutation-mat-simple-2d-cnn}{2}, \href{https://colab.research.google.com/drive/1oWuQg33bveHYqVET4IES_EWIH1nsa4yi}{3},\href{https://colab.research.google.com/drive/1oWuQg33bveHYqVET4IES_EWIH1nsa4yi}{4}.

{\bf Transformer.} We also performed experiments with a transformer model. However, it was able to solve only not-so-large groups like $n\le 15$. 
That model incorporated standard transformer components, including multi-head self-attention, positional encoding, feed-forward layers, and dropout to mitigate overfitting. However, the model struggled to converge to a satisfactory solution, even for relatively short sequences. This issue became increasingly severe as the sequence length grew. For instance, with $n=9$ (corresponding to path length = 107 and ideal = 36) or $n=13$ (path length = 332, ideal = 78), increasing the projection dimension to 512 or 1024, adding more layers, or training for more epochs failed to address the underlying challenges. Notebooks: \href{https://www.kaggle.com/code/nursmen/lrx-transformer-training}{1}, \href{https://www.kaggle.com/code/nursmen/lrx-transformer-use}{2}.

{\bf Transformer based on LRX specific features.}
The features inspired by Hamming distance were used:  difference between an element and its position;  difference with the left neighbor;  difference with the right neighbor.  A standard multihead attention with one encoder layer and MLP was used to estimate the number of steps required to reach the target state. The transformer demonstrated the ability to handle sequences up to $n=15$ without using beam search. For larger $n$, beam search is employed, and the maximum $n$ for which results were obtained is 100. The model and features only indirectly depend on $n$, therefore, when trained on micro batches constructed as random walks from the identity permutation for $n(n-1)/2$ moves, it can be used for different values of $n$, both larger and smaller.  (\href{https://www.kaggle.com/code/sergeifironov/permutation-transformer-solver-100}{Notebook}). 

{\bf Gradient boosting.} We also applied CatBoost and XGboost in our diffusion distance pipeline. Surprisingly, they perform not worse than neural networks (it was not the cases for Rubik's cube where boostings were not able to solve $4\times 4 \times 4$, $5\times5 \times5$ cubes), and even produce a bit shorter solutions. 
(\href{https://www.kaggle.com/code/antoninadolgorukova/lrx-gbm-powered-pathfinding}{Code}).

More simulation results for various parameters can be found in \href{https://www.kaggle.com/competitions/lrx-oeis-a-186783-brainstorm-math-conjecture/discussion/612338}{spreadsheet here}, or  on \href{https://docs.google.com/spreadsheets/d/1KFYTqPDSdH8_vdvWz3YLTLKg5IlW1Kkx_9kJClSKGOs/edit?usp=sharing}{google sheets}.

\section{Mathematical contributions}

In this section, we prove lower and upper bounds for the diameter and for particular elements, describe the decomposition of the conjecturally longest element of the desired length $n(n-1)/2$, describe algorithms to decompose in LRX generators, and formulate several conjectures on the growth, random walks and spectrum of the LRX Cayley graph. 

\subsection{Long and the longest element}

\begin{figure}[ht!]
   \centering
   \includegraphics[width=1.0\linewidth]{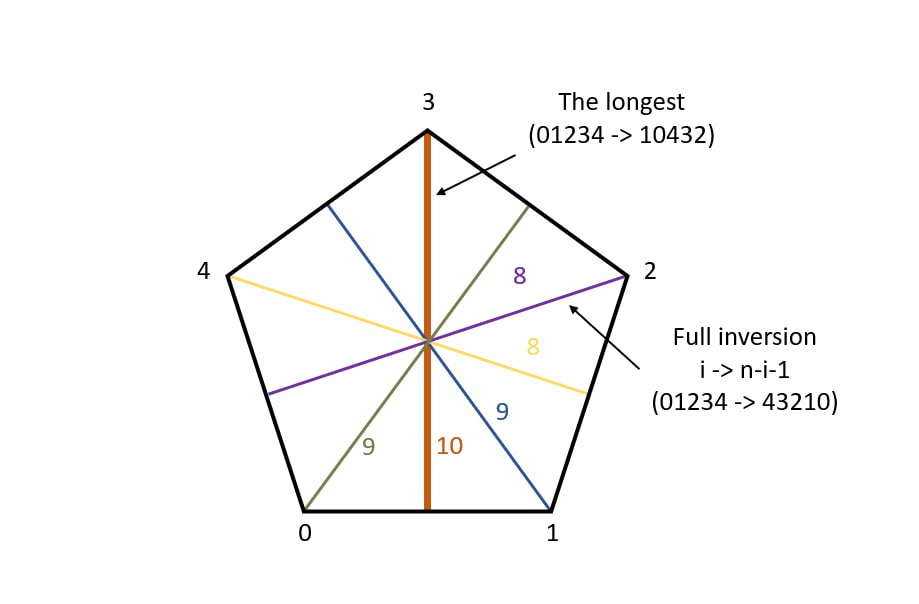}
   \caption{Permutations arising from dihedral symmetries are provably long elements. The longest is conjecturally one of them.}\label{fig:LongEl}
\end{figure}

We created an efficient implementation of the brute force breadth first search algorithm (see this \href{https://www.kaggle.com/code/ivankolt/lrx-4bit-uint64}{notebook}) using bit operations over \texttt{int64} and were able to traverse all the permutations for $n \le 15$, outperforming OEIS data by two steps (for $n=15$ computation took several days on Hadoop cluster). From that computation we observed that the longest element is unique and has a specific form~\eqref{eq:longest}. It can be described as a symmetry of a regular polygon (Fig.~\ref{fig:LongEl}), which is orthogonal to 0--1 edge. We conjecture that this element is the only longest element with length $n(n-1)/2$. 
Surprisingly, the full inversion $(n-1, n-2, \ldots, 2, 1, 0)$ is not the longest. As we will prove below, actually all the permutations which correspond to symmetries of a regular polygon are also long elements.  


\begin{proposition}
    Let 
    $$
    n > 2,\quad m = \Big\lfloor \frac n2 \Big\rfloor,\quad 
    \delta_n = [n = 2m] = \begin{cases}
        1, n \text{ \rm{is} even},\\
        0, n \text{ \rm{is} odd}.
    \end{cases}
    $$
    The element $\ell_n = (1\ 0 \ n-1 \ n - 2 \ldots 3 \ 2)$
    has the following explicit decomposition into product of LRX generators:
\begin{equation}
\label{eq:long}
\ell_n = \prod_{i=1}^{m-\delta_n} \big( X L^{(-1)^{i-1}} \big)^i \circ \prod_{i=m -1}^1 \big(L^{(-1)^{i-\delta_n}}X \big)^i \circ R^m.
\end{equation}    
\end{proposition}

The proof will be given elsewhere. The decompositions of that type were first found computationally: \href{https://www.kaggle.com/code/luxoove/lrx-optimal-algorithm-two-ways-bubble-sort}{notebook}.


\subsection{Lower bound on the diameter}
Below, we provide a lower bound on the diameter using a combinatorial argument. The key idea is that cyclic shifts preserve the cyclic order of any given triple of numbers. Therefore, for each triple where the order is disrupted, the move $X$ is required in the decomposition.   The symmetries of a regular polygon, as described above, correspond precisely to the permutations that maximize changes in cyclic order. (The code to generate these long elements: \href{https://www.kaggle.com/code/alexandervc/lrx-long-elements-2-f-petrov-oeisa186752}{notebook}.) 

\begin{proposition}
    The diameter of the LRX Cayley graph of $S_n$ is larger than or equal to $n(n-1)/2-n/2-1$.
\end{proposition}

First, we reformulate the setup as follows.

The numbers $1,\ldots,n$  are written in the vertices of a regular $n$-gon $A_1A_2\ldots A_n$, each number being written once. At each move, one can either rotate the arrangement by $2\pi/n$  counterclockwise or clockwise, or swap the numbers at $A_1$ and $A_2$.
For two arrangements $\pi_1,\pi_2$ denote by $d(\pi_1,\pi_2)$ the minimal number of moves which suffice to get $\pi_2$ from $\pi_1$.

\textbf{Claim.} If $\pi_2$ is obtained from $\pi_1$ by an axial symmetry, then $d(\pi_1,\pi_2)\geqslant n^2/2-n-1$.

\textbf{Proof.} Assume that $\pi_2$ is obtained from $\pi_1$ by several moves. Call two numbers $a, b\in \{1,\ldots,n\}$ friends, if they were swapped odd number of times. Since for every three numbers $a, b, c$ the orientation of a triangle formed by these numbers changed, there is odd number of pairs of friends between $a, b, c$. Let $A$ denote the set of all non-friends of element 1, $B$ the complement  of $A$, i. e. $B$ consists of 1 and its friends. Our condition yields that all elements of $A$ are mutual friends, and so are all elements of $B$, but there is no friendship between $A$ and $B$. Thus, the total number of pairs of friends is $n(n-1)/2-|A|\cdot |B|\geqslant n(n-1)/2-n^2/4$. Hence, there were at least as many swaps. Between any two swaps there should be a rotation (otherwise, we are doing something useless). Totally, the number of operations is not less than $n(n-1)-n^2/2-1=n^2/2-n-1$.

\subsection{Upper bound on the diameter and the algorithms}
We develop two algorithms which can decompose any permutation in the product of LRX generators. We prove  complexity bound $n(n-1)/2+3n$, thus bounding diameter from above. The code and tests for the first algorithm can be found in \href{https://www.kaggle.com/code/mixnota/article-project}{notebook}.

Its brief description is the following:

The algorithm decomposes a permutation into the product of cycles. After that, it works with each cycle \( (a_1, a_2, \dots, a_n) \) in the following way:

1) We compute the permutation corresponding to the cycle.

2) We initialize the special variable \( x \).

3) The element \( a_1 \) is placed in the \( a_n \)-th position, and we perform a sequence of elementary transpositions to move \( a_1 \) from the \( a_n \)-th position to the \( a_1 \)-th position. During each elementary transposition, we either increase or decrease \( x \) by 1. The sign of the change depends on the direction of the elementary transposition. Essentially, \( x \) represents the current position of the transposition, but if our transposition is \( (1, N) \), the change in \( x \) is also \( \pm1 \), even though the position changes from 1 to \( N \) or from \( N \) to 1.

4) Next, we attempt to restore the position of \( a_2 \) and compute the sequence of elementary transpositions that move \( a_2 \) to the \( a_2 \)-th position. We apply the same operations to \( x \), and so on.

5) We repeat these actions for every element of the cycle.

6) In certain situations, the elementary transposition should be replaced by \( L \) or \( R \) (depending on the direction in which we are moving the current element). This replacement should occur if and only if the variable \( x \) satisfies the condition:

\[
x - c = \pm (N - 1)
\]

\textbf{Complexity estimation.}
First, when decomposing a cycle into transpositions, we do not shift anything by more than 1, except for elements already affected by the cycle.
Second, when moving an element from position \( i \) (or \( i \pm 1 \)) to position \( \sigma(i) \), we perform a sequence of at most \( \text{dist}(i, \sigma(i)) + 1 \) transpositions.
Third, the transition from \( i \to \sigma(i) \) to \( \sigma(i) \to \sigma(\sigma(i)) \) takes at most one rotation.
Thus, we perform at most  
\[
\sum 2(\text{dist}(i, \sigma(i)) + 1) + (\text{number of elements in the cycle})
\]
operations. The transitions between cycles and returning back will take at most \( n \) rotations.
Therefore, the entire decomposition takes no more than  
\[
2n + \sum 2(\text{dist}(i, \sigma(i)))
\]
operations.

Finally, after minimization,
\[
\sum \text{dist}(i, \sigma(i)) \leq \frac{n^2}{4}
\]
which gives a total of  
\[
2n + \frac{n^2}{2} + \frac{n}{2}
\]
operations.

We also develop a second algorithm: \href{https://www.kaggle.com/code/luxoove/top1-lrx-inversions-based-algorithm}{notebook}. Currently, it is our top performing algorithm from the practical tests, we expect its complexity is bounded by $n(n-1)/2+n/2$, however, it is not yet proved.

It might be expected that optimal algorithm with polynomial complexity can be developed, but it is not achieved yet. 

\subsection{Conjectures: Gumbel for growth, spectrum uniformity, random walks mixing, etc. }

Based on explicit computations for the growth for $n \le 15$ and its analysis (notebooks \href{https://www.kaggle.com/code/ogurtsov/gumbel}{1} \href{https://www.kaggle.com/code/ogurtsov/gumbel-for-binary-puzzle}{2} ) we come to the following conjecture, which can be thought as an analogue of the central limit theorems. It is in the vein of works by P.Diaconis et.al. ~\cite{diaconis1977spearman, diaconis1988metrics,chatterjee2017central} where for Coxeter generators growth and some other statistics were shown to follow Gaussian normal law in a limit. A new point of our conjecture is the appearance of asymmetric Gumbel distribution. Such asymmetry is quite natural since for the randomly chosen generators growth will be highly asymmetric with exponential growth and sharp abrupt (\href{https://mathoverflow.net/q/322877/10446}{Rubik's cube growth} is a prototypical example), while for nearly commutative Coxeter generators the growth is symmetric Gaussian. A slight asymmetry of growth for LRX generators indicates its intermediate nature between "exponential growth" and "commutative-like Gaussian growth" cases.

The growth distribution of the LRX graph seems to follow the Gumbel distribution for large $n$. The detailed analysis is provided in supplementary section~\ref{growth}.

\begin{figure}[ht!]
   \centering
   \includegraphics[width=1.0\linewidth]{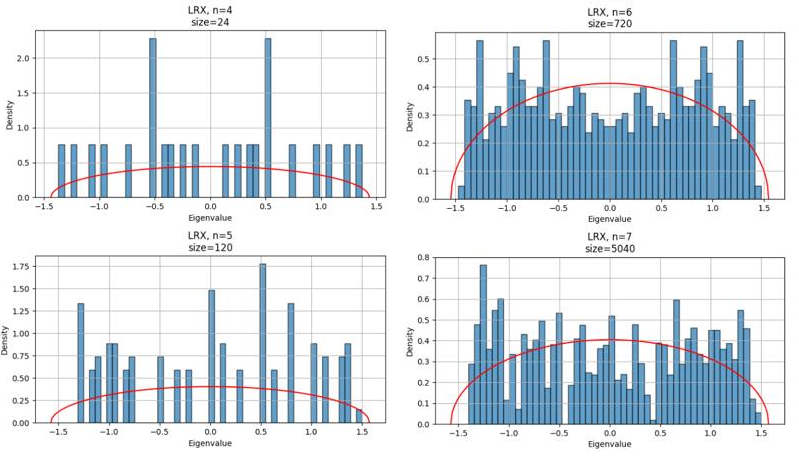}
   \caption{Spectrum distribution of the LRX graphs. }\label{fig:LRX-spec}
\end{figure}

The analysis of the spectrum of LRX graphs is performed in \href{https://www.kaggle.com/code/nikolenkosergei/spectrum-analysis}{notebook}. From the figure above it is natural to expect that spectrum tends to uniform distribution.

\begin{figure}[ht!]
   \centering
   \includegraphics[width=1.0\linewidth]{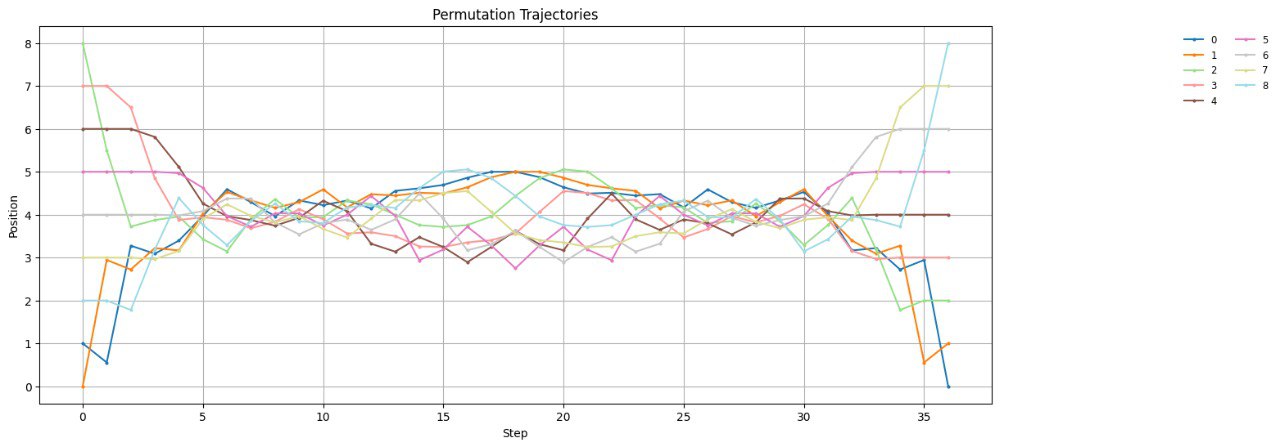}
   \caption{Analogs of "random sorting networks" for LRX. Trajectories of individual elements under average of all possible shortest trajectories between longest node and identity. }\label{fig:LRX-sort}
\end{figure}

We have computed all possible shortest paths from the conjecturally longest element to the identity by methods of the dynamical programming (\href{https://www.kaggle.com/code/fedmug/lrx-longest-paths}{notebook} . And after that analyzed averaged trajectories (\href{https://www.kaggle.com/code/antoninadolgorukova/lrx-sorting-networks}{notebook}) which are analogs of random sorting networks~\cite{angel2007random}. The figure \ref{fig:LRX-sort} below represents the results. It seems the pattern is different from the "sine curves" discovered in~\cite{angel2007random}. 

\href{https://www.kaggle.com/code/ogurtsov/gumbel-for-binary-puzzle}{notebook}. 

\section{Code availability}

The code is available at the Kaggle platform, where it can be easily launched:
\href{https://www.kaggle.com/competitions/lrx-oeis-a-186783-brainstorm-math-conjecture/code}{LRX OEIS-A186783 brainstorm math conjecture}

The link above leads to one of the three public Kaggle challenges which we created to stimulate the research and interplay between artificial intelligence and mathematical communities, the other two:
\href{https://www.kaggle.com/competitions/lrx-binary-in-search-of-gods-number}{LRX discover math and God's algorithm}, \href{https://www.kaggle.com/competitions/lrx-discover-math-gods-algorithm}{LRX-binary: in search of God's number}. Kaggle infrastructure provides a convenient way to benchmark and compare different algorithms by just making the submissions to these challenges. The code, the data and discussions are stored at the same place. The code can be executed for free on Kaggle cloud servers. The first challenge is to decompose the conjecturally longest elements and to find their shortest decompositions. The participants have already achieved the length $n(n-1)/2$; thus, if we believe the conjecture is true, the optimum is found. The second challenge is to decompose random permutations and aims to find optimal algorithms (at the moment of writing optimum is not achieved). The third one asks the same but for the binary strings acted by LRX generators.

\subsubsection*{Acknowledgements}
A.C. is deeply grateful to M. Douglas for his interest in this work, engaging discussions, and invitation to present preliminary results at the Harvard CMSA program on "Mathematics and Machine Learning" in Fall 2024; to M. Gromov, S. Nechaev, and V. Rubtsov for their invitation to give a talk (\href{https://youtu.be/RkmBwlSyhfA?si=KgqQtRFaqx5ykd_s}{video}) at "Representations, Probability, and Beyond: A Journey into Anatoly Vershik World" at IHES, as well as for stimulating discussions. A.C. is grateful to J. Mitchel for involving into the Kaggle Santa 2023 challenge, from which this project originated, and to M.Kontsevich, Y.Soibelman,  S.Gukov, A. Hayat, T. Smirnova-Nagnibeda,  D.Osipov, V. Kleptsyn, G.Olshanskii, A.Ershler, J. Ellenberg, G. Williamson, A. Sutherland,  Y. Fregier, P.A. Melies, I. Vlassopoulos, F.Khafizov, A.Zinovyev, H.Isa-mbert, T. Rokicki for the discussions, interest and comments, to his wife A.Chervova and daugther K.Chervova for support, understanding and help with computational experiments.  

We are deeply grateful to the many colleagues who have contributed to the CayleyPy project at various stages of its development, including: N.Bukhal, J.Naghiev, A.Lenin, E.Uryvanov,  A. Abramov, M.Urakov, A.Ku-chin,  B.Bulatov,  F.Faizullin, A.Aparnev, O.Nikitina, A.Titarenko, U.Kniazi-uk, D.Naumov, A.Krasnyi, S.Botman, A.Kostin,
R.Vinogradov, I.Gaiur, 
K.Y-akovlev, V.Shitov, E.Durymanov, A.Kostin, R.Magdiev, M.Krinitskiy, P.Sn-opov.

\section{Supplementary}

Here we present additional analysis related both to mathematical and AI counterparts presented in the main text.
Subsection \ref{growth} we give details on the growth approximation by Gumbel distribution, the next subsection devoted to similar analysis
for the Schreier coset graph and formulating conjectures for it.
Subsection \ref{sec-mix} we analyze X-trick and random walks, in particular we conjecture that mixing time for random walks with X-trick is $O(n^2)$,
that means X-trick speed-ups mixing comparing to standard random walks which mixing time is known to be between $O(n^3)$ and $ O(n^3log(n))$. 
Which partially explains why X-trick is so effective in beam search.
Subsection \ref{sec-dif-dist} gives details on the diffusion distance and provides results confirming that it is not fully perfect heuristics, which justifies our main algorithmic proposal - combining it with RL methods.
Subsection \ref{sec-beamsize-and-depth} presents analysis how performance depends on the beam width and depth of non-backtracking. 
(More simulation results for various parameters can be found in \href{https://www.kaggle.com/competitions/lrx-oeis-a-186783-brainstorm-math-conjecture/discussion/612338}{spreadsheet here}, or  on \href{https://docs.google.com/spreadsheets/d/1KFYTqPDSdH8_vdvWz3YLTLKg5IlW1Kkx_9kJClSKGOs/edit?usp=sharing}{google sheets}.)
In subsection \ref{sec-tropical} we discuss relation of the tropical mathematics and the Bellman equation.  

\subsection{Gumbel growth distribution}
\label{growth}

A \textit{layer} of a Cayley graph is the set of all vertices with the same distance to the target (e.g. identity permutation $e$). LRX Cayley graph for $S_n$ conjecturally consists of $\frac{n(n-1)}2 + 1$ layers. Using the BFS algorithm, we calculate the sizes of all these layers for $n\le 15$. Being normalized, the layer sizes form the \textit{growth distribution} of the graph. We hypothesize that for $n \to \infty$ it converges to the left-skewed Gumbel distribution with pdf:

\begin{equation}
\label{eq:gumbel}
    p(x \vert \mu, \beta) = \frac 1\beta F\Big(\frac{x-\mu}\beta \Big), \quad F(t) = e^{t - e^t}.
\end{equation}

\begin{figure}[ht!]
   \hspace*{-13mm}
   \includegraphics[width=1.2\linewidth]{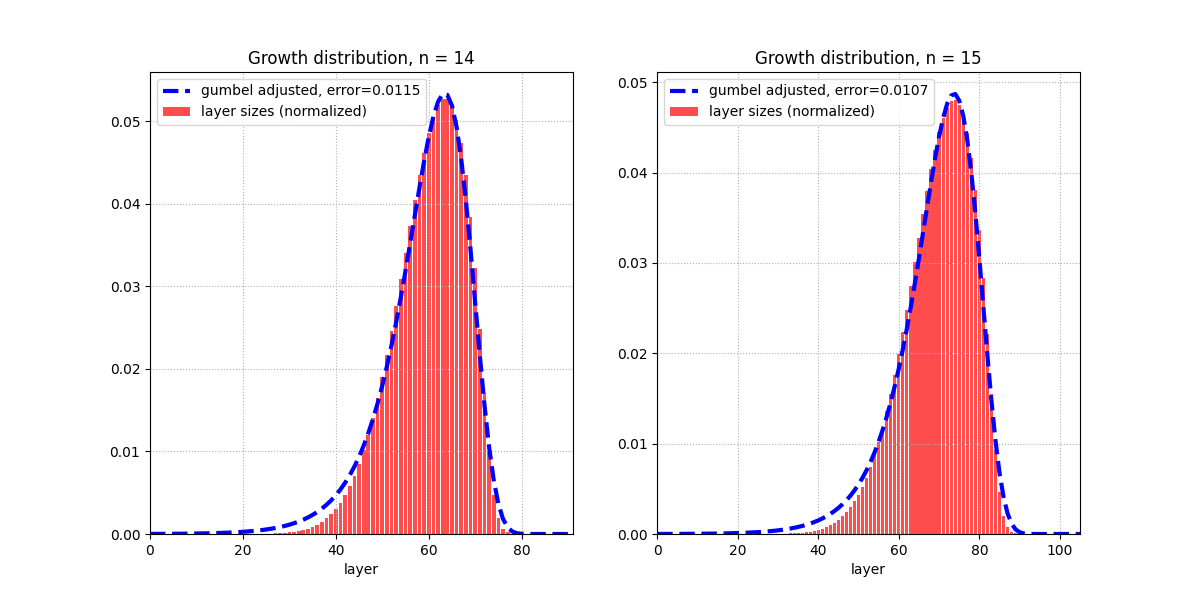}
   \caption{Growth distribution of LRX Cayley graph fitted by left-skewed Gumbel~\eqref{eq:gumbel}.}\label{fig:LRX-growth}
\end{figure}

\begin{figure}[ht!]
   \hspace*{-15mm}
   \includegraphics[width=1.2\linewidth]{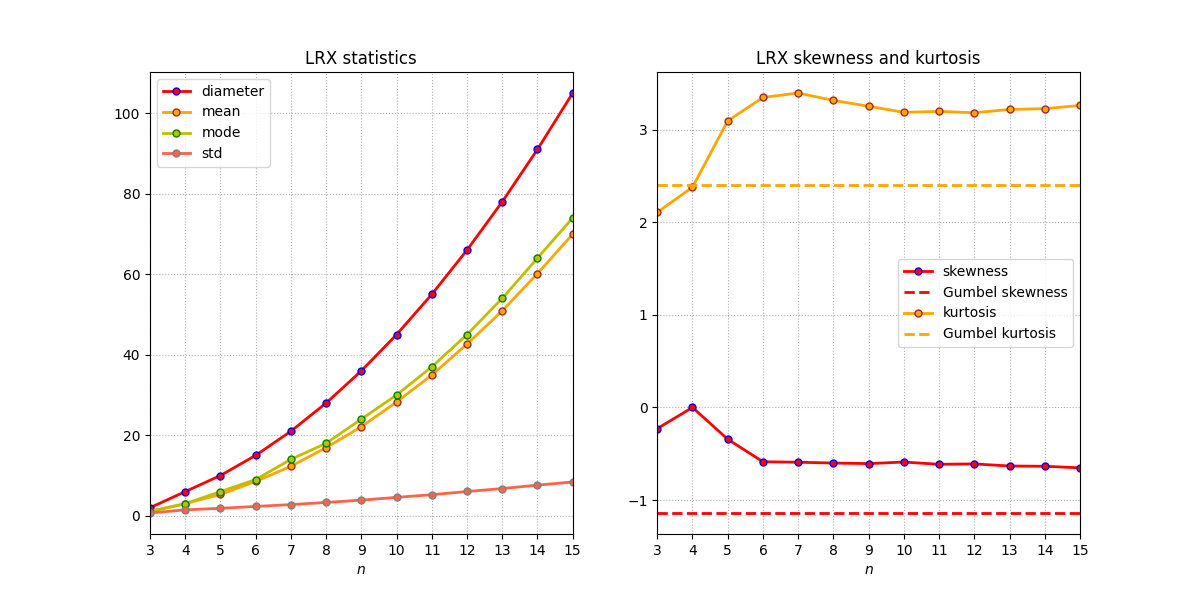}
   \caption{LRX growth statistics. Diameter, mean, mode, $\mathrm{std}^{3/2}$ seem to be quadratic polynomials of $n$. Skewness is a slightly negative monotonic function of $n$.}\label{fig:LRX-stats}
\end{figure}

To estimate the parameters $\mu$ and $\beta$, we calculate the mean and variance of the LRX growth distribution and equal them to $\mu - \gamma \beta$ and $\frac{\pi^2}6 \beta^2$, respectively (theoretical moments of~\eqref{eq:gumbel}). After that, we search by grid for slightly better parameters. The approximation obtained for the growth of LRX by the Gumbel distribution is presented in Fig.~\ref{fig:LRX-growth}. 

The mean, mode and standard deviation raised to the power of $\frac 32$ seem to be quadratic functions of $n$ (see Fig.~\ref{fig:LRX-stats}). The comparison of these LRX statistics with those obtained by the Gumbel fit is presented in Table~\ref{tab:growth_stats}. In all cases we observe a good fit by quadratic polynomials with higher coefficients within $[0.1, 0.5]$. See \href{https://www.kaggle.com/code/fedmug/lrx-growth-distribution}{notebook} for details.

The skewness and excess kurtosis are shown in the right part of Fig.~\ref{fig:LRX-stats}. Although they are not very close to the theoretical values of the Gumbel distribution, the skewness, not converging to $0$, reveals that the limit growth distribution is not Gaussian. 
\clearpage

\begin{table}[t]
    \centering
    \begin{tabular}{|c|l|l|}
        \hline
        \textbf{Statistic} & \textbf{LRX growth} & \textbf{Gumbel fit} \\
        \hline
        diameter & $0.5n^2 - 0.5n$ & ---\\
        mean & $0.3763n^2 - 1.0698n + 1.3015$ & $0.3765n^2 - 1.1205n + 1.3647$\\
        mode & $0.3896 n^2 - 0.983n + 1.0335$ & $0.3858 n^2 - 0.9545n + 1.1767$ \\
        $\mathrm{sd}^{3/2}$ & $0.1455n^2 - 0.7245n + 2.4859$ & $0.1686 n^2 - 0.6196n + 1.6563$ \\
        \hline
    \end{tabular}
    \caption{Quadratic fits for statistics of LRX growth.}
    \label{tab:growth_stats}
\end{table}

\subsection{Coset graph}
\begin{figure}[ht!]
   \hspace*{-13mm}
   \includegraphics[width=1.2\linewidth]{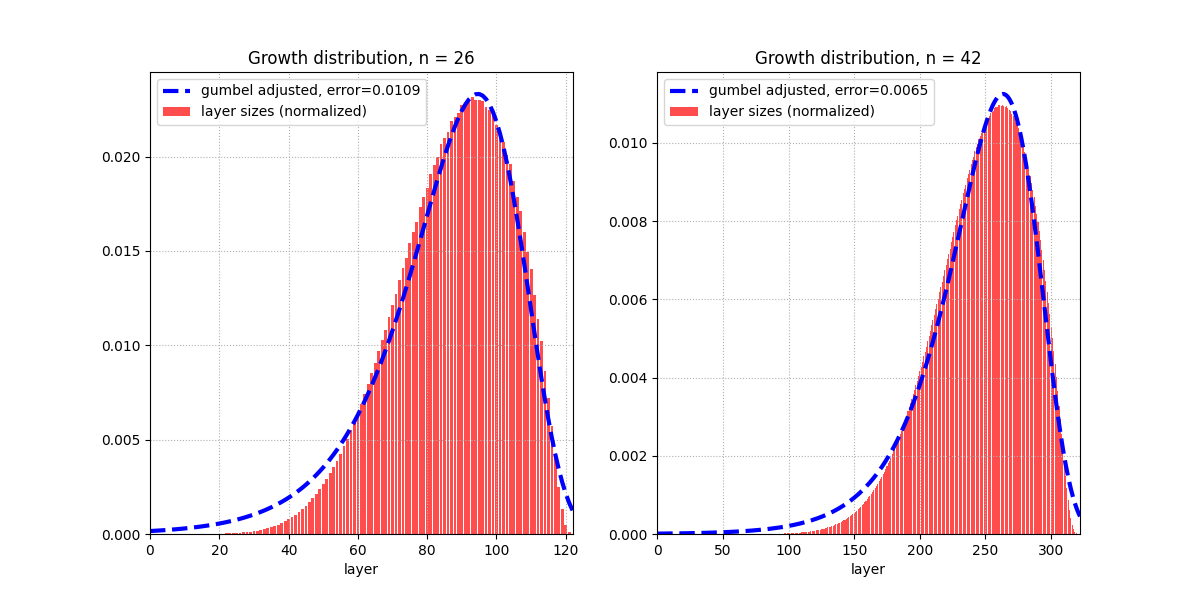}
   \caption{Growth distribution of the coset graph fitted by left-skewed Gumbel~\eqref{eq:gumbel}.}\label{fig:coset-growth}
\end{figure}

A similar analysis is performed for the Schreier coset graph for $S_{n}, n=2m$ on binary strings with $m$ zeros and $m$ ones (analogue of $Gr(m,2m)$ over the "field with one element"), $4\le m \le 21$.  Efficient \href{https://www.kaggle.com/code/ivankolt/lrx-4bit-uint64-coset}{code} that stores such strings in \texttt{int64} allowed us to traverse the graph up to $n=42$ and compute its growth. Once again, 
 it can be well approximated by the Gumbel distribution~\eqref{eq:gumbel} (Fig.~\ref{fig:coset-growth}).
Note: in case of Schreier graphs growth is dependent on the starting point in general - we take it to be first $m$ zeros, than $m$ ones.
In contrast to Cayley graphs we will speak not about the diameters of these graphs, but the about distance to the most far nodes from that
starting point - we will call them God's numbers - using terminology adopted in puzzle community.

\begin{figure}[ht!]
   \hspace*{-15mm}
   \includegraphics[width=1.2\linewidth]{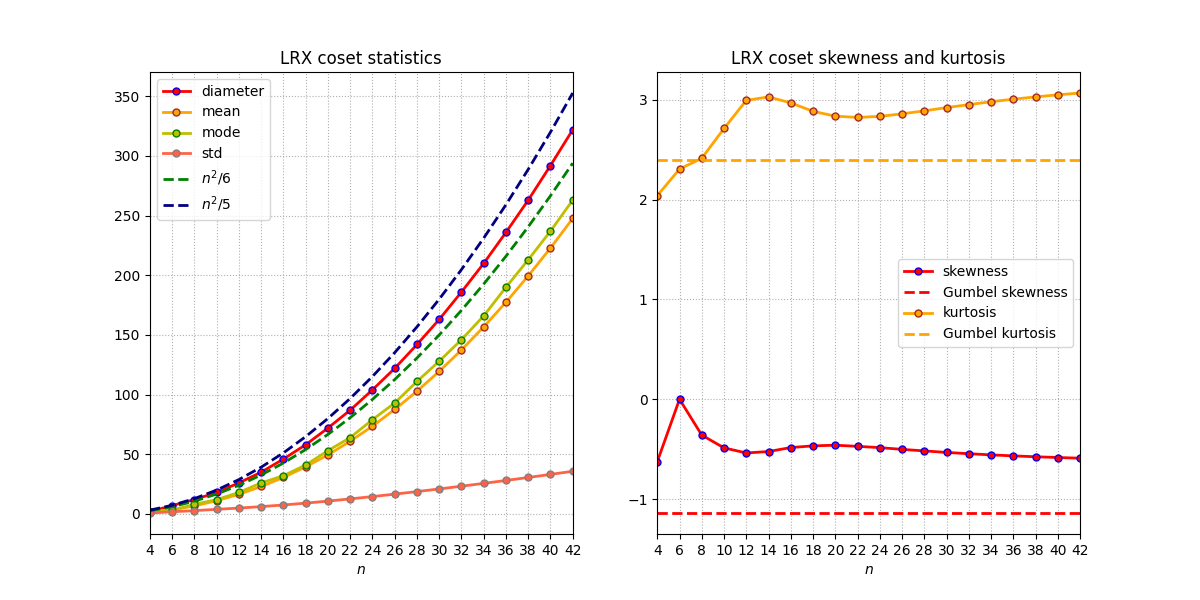}
   \caption{Coset growth statistics.}\label{fig:coset-stats}
\end{figure}

{\bf Conjectures.}
The growth  coset growth converges to the Gumbel~\eqref{eq:gumbel} in distribution for $n\to \infty$. 
The coset growth statistics for mean, mode can be approximately described by quadratic as presented in Fig.~\ref{fig:coset-stats} and Table.~\ref{tab:coset_growth_stats}, respectively.
Moreover it is tempting to conjecture that the God's numbers $d_n$ is (which surprisingly coincide with 
\href{https://oeis.org/A132297}{A132297}( Number of distinct Markov type classes of order 2 possible in binary strings of length n)
and with \href{https://oeis.org/A267529}{A267529}(Total number of ON (black) cells after n iterations of the "Rule 141" elementary cellular automaton starting with a single ON (black) cell)\footnote{We thank Tomas Rokicki for that observations}: 
$$
    d_n = \frac 3{16}n^2 - \frac n4 + 2 - \frac{n \mod 4}8, n \geq 6, n \text{ is even},
$$

See  (\href{https://www.kaggle.com/code/fedmug/lrx-coset-growth}{notebook}) for more details.
(God's number is quasi-polynomial - which is consistent which general conjecture presented in the subsequent paper:
\cite{CayleyPyGrowth}).

 \begin{table}[ht!]
    \centering
    \resizebox{0.9\textwidth}{!}{
    \begin{tabular}{|c|l|l|}
        \hline
        \textbf{Statistic} & \textbf{Coset growth} & \textbf{Gumbel fit} \\
        \hline
        God's number & $0.1875n^2 - 0.25n + O(1)$ & ---\\
        mean & $0.1591 n^2 - 0.8887n + 3.4809$ & $0.1595n^2 - 0.9919n + 3.9014$\\
        mode & $0.169n^2 - 0.936n + 3.8898$ & $0.1657n^2 - 0.7831n + 3.2604$ \\
        $\mathrm{sd}^{3/2}$ & $0.1597 n^2 - 1.8523n + 8.3784$ & $0.1927n^2 - 1.757n + 6.6497$ \\
        \hline
    \end{tabular}}
    \caption{Quadratic fits for statistics of the Schreier coset growth.}
    \label{tab:coset_growth_stats}
\end{table}

{\bf Long elements.} Currently we are unable to find a clear pattern for the longest elements, 
however we are able to identify some elements which are quite long.
The sizes of the last layer of the coset graph, which contain the longest elements, seem to have no clear pattern (Table~\ref{tab:coset_longest}). The projection of the conjecturally longest element~\eqref{eq:longest} of the LRX Cayley graph onto the binary coset by the rule $ \big[\ell_n^i < \frac n2\big]$, $0 \leq i <n$, produces
$00\underbrace{11\ldots 1}_{\lceil \frac n2 \rceil}\underbrace{00\ldots 0}_{\lfloor \frac n2 \rfloor - 2}
$,
whose length obviously does not exceed $\frac n2 - 2$ and hence these elements (longest in the full Cayley graph) cannot be longest in the coset graph. We propose a family of long elements whose length for feasible $n$ is close to the God's numbers $d_n$:

\begin{equation}
    c_n = \underbrace{11\ldots 1}_{n_0}\underbrace{00\ldots 0}_{n_1} \underbrace{11\ldots 1}_{n_2}\underbrace{00\ldots 0}_{n_3}, \quad n_i = \Big\lfloor \frac{n+i}4\Big\rfloor.
\end{equation}

We examine these elements and all their cyclic shifts, and calculate the distance to the central element $00\ldots 011\ldots 1$. The maxima of these distances, presented in the last column of Table~\ref{tab:coset_longest}, turn out to be close to the diameter $d_n$.

{\bf Conjecture.} The lengths of these elements are not smaller than God's number minus $O(n)$.  

\begin{table}[ht!]
\centering
\resizebox{0.9\textwidth}{!}{
\begin{tabular}{|c|c|c|c|c|c|}
\hline
$n$ & \#Longest & God's numbers $d_n$ & $\mathrm{len}(c_n)$ & $\max\limits_{0\leq k < n}\mathrm{len}(L^kc_n)$ & $\max\limits_{s \leftrightsquigarrow c_n} \mathrm{len}(s)$ \\
\hline\hline
4  & 3  & 2  & 2 & 2 & 2 \\ \hline
6  & 1  & 7  & 3 & 6 & 7  \\ \hline
8  & 1  & 12 & 10 & 10 & 10 \\ \hline
10 & 4  & 18 & 13 & 17 & 18 \\ \hline
12 & 4  & 26 & 23 & 24 & 24 \\ \hline
14 & 11 & 35 & 27 & 32 & 34 \\ \hline
16 & 6  & 46 & 42 & 43 & 43 \\ \hline
18 & 14 & 58 & 47 & 53 & 57 \\ \hline
20 & 10 & 72 & 67 & 69 & 69 \\ \hline
22 & 32 & 87 & 73 & 80 & 84 \\ \hline
24 & 16 & 104 & 98 & 100 & 100 \\ \hline
26 & 49 & 122 & 105 & 113 & 119 \\ \hline
28 & 26 & 142 & 135 & 138 & 138 \\ \hline
30 & 99 & 163 & 141 & 152 & 158 \\ \hline
32 & 42 & 186 & 178 & 181 & 181 \\
\hline
\end{tabular}}

\caption{Number of longest elements, the length of the long (but not the longest) elements $c_n$, maximal length of their shifts and the length of the longest element reachable from these shifts by distance non decreasing steps  for coset graphs. The values in the last three columns seem to be as close to $d_n$ as $O(n)$.}
\label{tab:coset_longest}
\end{table}

\subsection{Mixing time}\label{sec-mix}

P.~Diacons and L.~Soloff-Coste studied random walks on LRX Cayley graphs \cite{diaconis1993comparison}. 
They estimated the \href{https://en.wikipedia.org/wiki/Markov_chain_mixing_time}{\textit{mixing time}} --- the number of steps needed for a permutation to resemble a uniform random sample --- to be between
$n^3$ and $n^3\log n$.
Here we complement their fundamental results by computational analysis of the different types of the random walks on the same LRX Cayley graph
and formulate new conjectures (items 2,3): 

\begin{enumerate}
\item 
pure random walks (all moves L, R, X are equiprobable on each step) - mixing time is between $n^3$ and $n^3\log n$~\cite{diaconis1993comparison}
\item 
random walks with non-backtracking (forbids going back where it just came from) - conjecture: mixing time has the same order  between $n^3$ and $n^3\log n$
\item 
random walks with non-backtracking and X-trick -  conjecture: mixing time reduces to  $n^2$ 
\end{enumerate}

\begin{figure}[ht!]
   \centering
   \includegraphics[width=1.0\linewidth]{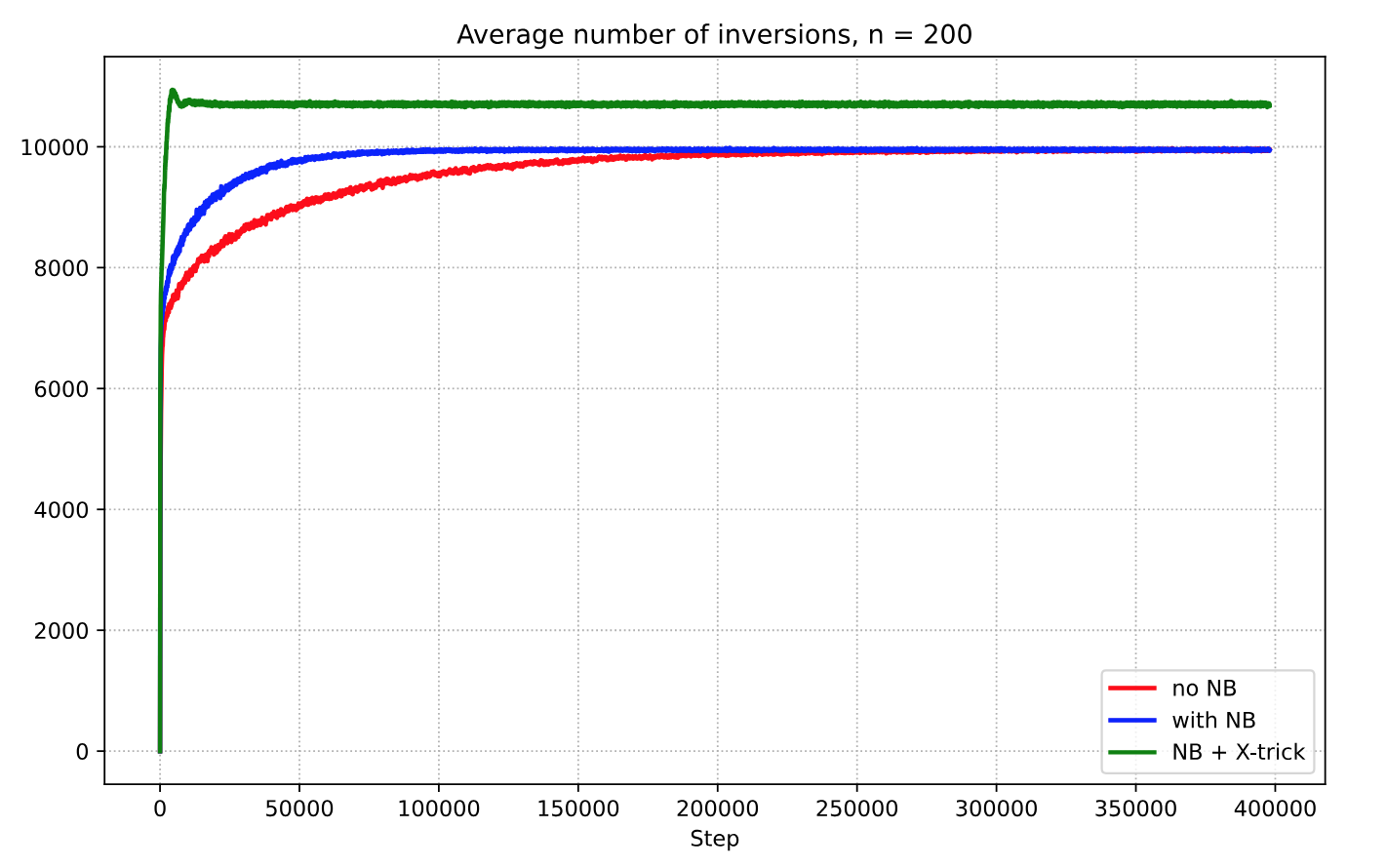}
   \caption{Pure random walks, random walks with non-backtracking, and random walks with non-backtracking and X-trick. The latter mixes faster than the first two.}\label{fig:LRX-mixing}
\end{figure}

In order to analyze the behavior of the random walks we consider the following statistics - number of inversions.
The advantage of that statistics that it can be computed very fast. 
Figure \ref{fig:LRX-mixing} presents values of that statistics averaged over  $5000$ random walks.
Roughly speaking the time (step) of achieving the plateau corresponds to the mixing time,
more precisely such time is lower bound for the mixing time. 
(As expected for the first two series the level of the plateau is $\frac{n(n-1)}4$ which coincides with expectation of the number of inversions of a random permutation, while for X-tricks random walks it is different). 

To estimate the mixing time we used the step when a series of random walks reaches 95\% or 99\% of the asymptote (\href{https://www.kaggle.com/code/fedmug/lrx-random-walk-asymptote}{notebook}). For random walks of type~(1) and~(2) these time steps are fitted by a cubic polynomial - that corresponds to mixing time being   between $n^3$ and $n^3\log n$.  Interestingly, the best fit for the series~(3) is delivered by a quadratic polynomial. Consequently, our conjecture for the mixing time is $O(n^3)$; introducing X-trick allows to mix faster in $O(n^2)$, however, the distribution in this case is clearly not uniform. Such speed-up of mixing may serve as an explanation why X-trick is so effective in beam-search.

\subsection{LRX diffusion distance analysis}\label{sec-dif-dist}

Diffusion distance is a very important part of our AI framework. 
Roughly speaking it should be thought as a length of a random path between two nodes. 
That phrasing is almost correct, but hides the  detail which is quite important. There is actually no "THE" diffusion distance,
there is no way to define one canonical diffusion distance on a graph, it is always a family depending on some parameter. 
A conceptual explanation: 
in a sense diffusion distance is a deformation of the canonical distance (length of the shortest path),
but since it is a deformation it must depend on some parameter. 
The most straight forward way to introduce the parameter is the following:
we fix some number of steps $K_{max}$ and consider random paths of the length $K_{max}$, then the average number of steps,
which such paths needs to get from node $A$ to node $B$ corresponds to the diffusion distance. 
Such way is similar to \href{https://en.wikipedia.org/wiki/Brownian_bridge}{Brownian bridge}  -  one needs to fix time $T$ to define it - that is similar to our fixation of the $K_{max}$.  

{\bf Definition of the diffusion distance. } Consider a graph, let $S$ be some selected node (e.g. "solved state", identity in group). 

Let us denote by $P(V,K)$ the transition probability from node $S$ to node $V$ in $K$ steps. I.e.: 
\begin{equation}
    P(V,K) = \text{Probability to reach V from S in K steps }
    \label{eq:P}
\end{equation}
That is -- one considers all possible paths of length $K$ and starting from $S$. And one divides number of such paths ending at $V$, by the number all such paths. (In particular we may get zero if $V$ is not reachable it $K$ steps.)

 Diffusion distance to node $V$ from node $S$ for the random walks length fixed to be  $K_{max}$ can be defined:
\begin{equation}
    \mathrm{DD}(V,K_{max}) = \frac{\sum_{k=0,...,K_{max} } ~  k ~  P(V,k)  }{\sum_{k=0,...,K_{max} }  P(V,k)}
\end{equation}
The sense of the formula above is clear -  it is weighted average of numbers of steps of random walks: $1,2,3,...K_{max}$ where weights are given by probabilities to reach $V$ in these numbers of steps. So it is exactly "average length of a random path from S to V" in certain precise sense and dependence on $K_{max}$ is explicit. The numerator and denominator are different only by presence  of $k$ in numerator.

{\bf Formula for the diffusion distance. } (Case of simple random works - without non-backtracking or any other modification). 
Diffusion distance 
can be computed via the adjacency matrix of a graph as follows: 

\begin{equation}
    \label{eq:dd}
    \mathrm{DD}(V,K_{max} ) = \frac{\sum\limits_{k=0}^{K_{\max}} k  P(V, k)}{\sum\limits_{k=0}^{K_{\max}}   P(V, k)}
\end{equation}
    
\noindent
where $\mathbb P(v, k)$ is the probability to reach $v$ from the initial state $e$ in $k$ steps. If $A$ is the adjacency matrix of the Cayley graph then $\mathbb P(v, k) = \frac{A^k_{0v}}{\sum\limits_j A^k_{0j}}$. 
The formula reveals a striking similarity of diffusion distance to zeta functions, and supersymmetric Landau-Ginzburg models. We will discuss it in subsequent publications. 

\begin{figure}[ht!]
\centering
\includegraphics[width=1.0\linewidth]{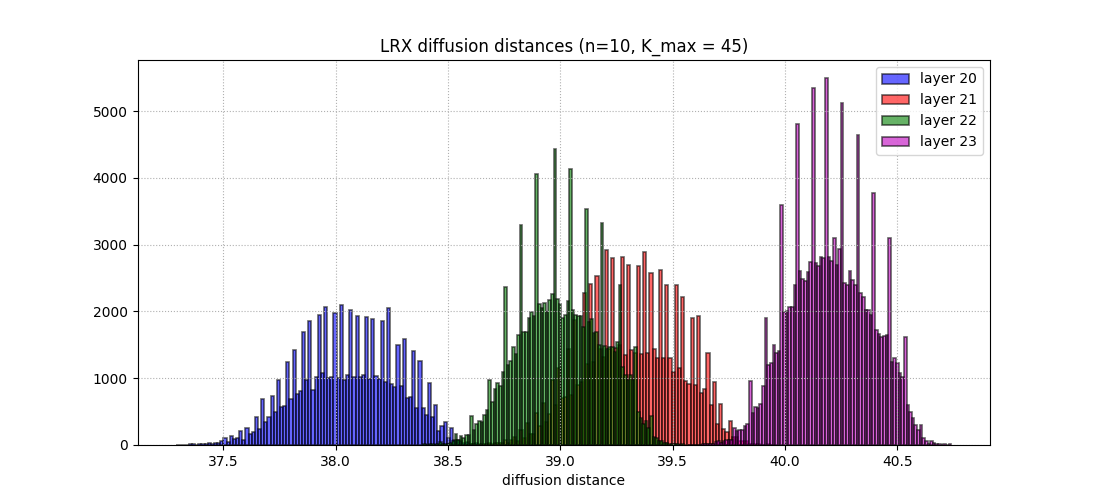}
\caption{Histogram of diffusion distances for layers $20$--$23$ of LRX Cayley graph for $S_{10}$. The average diffusion distance for layer $22$ is clearly less than for layer $21$ - demonstrating non-monotonicity of the diffusion distance with respect to the actual distance. }\label{fig:LRX-dd-10}
\end{figure}

{\bf Computations and analysis. } 
Calculation of powers of the adjacency matrix could be infeasible even for relatively small $n$. To speed up the computations, we use dynamic programming to calculate $A^k_{0v}$ --- number of paths of length $k$ from the identity permutation $e$ to the state $v$.
Results of the computations presented on the figure \ref{fig:LRX-dd-10}.
Conclusions: first diffusion distance has a huge variation when the actual distance (length of the shortest path is fixed),
one can see that histograms of the diffusion distance for layers $k$, $k+1$ intersect - which is unpleasant feature for diffusion
distance utilization in path-finding, since it will mislead  not to approach the goal, but deviate from it in some cases.
Second moreover, in some cases the same happens  not only for exceptional cases, but even in average diffusion distance 
for e.g. layer $22$ is smaller than for the layer $21$. 
These are the features demonstrating that diffusion distance is a not a perfect predictor of the true distance,
however from practical experience we see that it is working quite well in many cases, but still RL approach presented in the present paper
might be more promising.

\subsection{Impact of beam width and non-backtracking on the beam search performance}\label{sec-beamsize-and-depth}

The key parameter of beam search is  beam width (beam size). But advanced versions introduce various other parameters,
in particular non-backtracking depth, which controls how many previous moves are forbidden from being immediately revisited (to avoid short cycles). Varying these parameters affects performance: both the success rate and the length of the solution paths. Here describe results of such analysis. 

To study how beam width and non-backtracking depth affect the performance of ML-guided beam search, we conducted over 15,000 experiments (30–50 runs per condition) using XGBoost, the fastest model to date. We varied the permutation size ($n$), the beam width, and the non-backtracking depth (\href{https://www.kaggle.com/code/antoninadolgorukova/lrx-rw-bs-parameters-in-gbdt-guided-beam-search}{kaggle notebook}).
We tested all approaches on the element of the conjecturally longest element of length $n(n-1)/2$ described in previous sections.
These experiments revealed that increasing both parameters consistently improves the success rate, though gains diminish for larger $n$. 

\begin{figure}[H]
  \centering
  \includegraphics[width=0.8\textwidth]{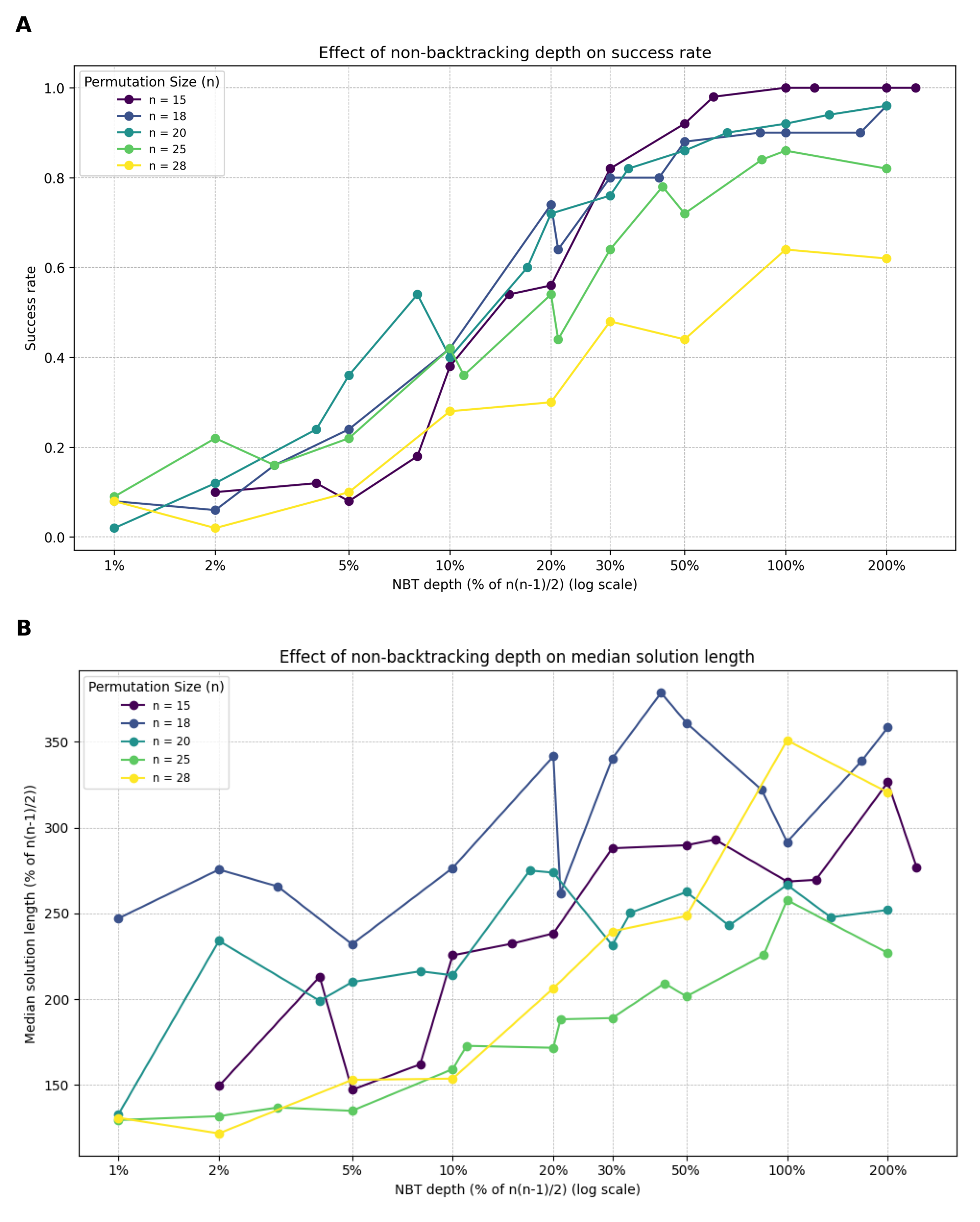}
  \caption{Effect of non-backtracking (NBT) depth for beam search on (A) success rate and (B) median solution length.
  Beam width is fixed at the minimum viable value for each permutation size $n$.}
  \label{fig:nbt_xgboost}
\end{figure}

\enlargethispage{3em}
To isolate the effect of non-backtracking, we set the beam width to the minimum value that yielded a non-zero success rate for each $n$ (15, 18, 20, 25, and 28) and systematically increased the depth of non-backtracking. As shown in Fig. \ref{fig:nbt_xgboost}A, increasing the depth of non-backtracking (expressed as a percentage of the conjectured optimal solution length, $n(n-1)/2)$) improves the success rate, although the magnitude of the improvement is smaller for larger permutations. In addition, deeper non-backtracking typically results in longer median solution lengths (Fig. \ref{fig:nbt_xgboost}B).
However, one cannot directly interpret that phenomena as decreasing performance, 
since success rate is improving - that beam search is able to solve cases more difficult for it,
that difficulty  manifest itself in  larger lengths and that might be the reason why  median solution length is growing. 

\begin{figure}[H]
  \centering
  \includegraphics[width=\textwidth]{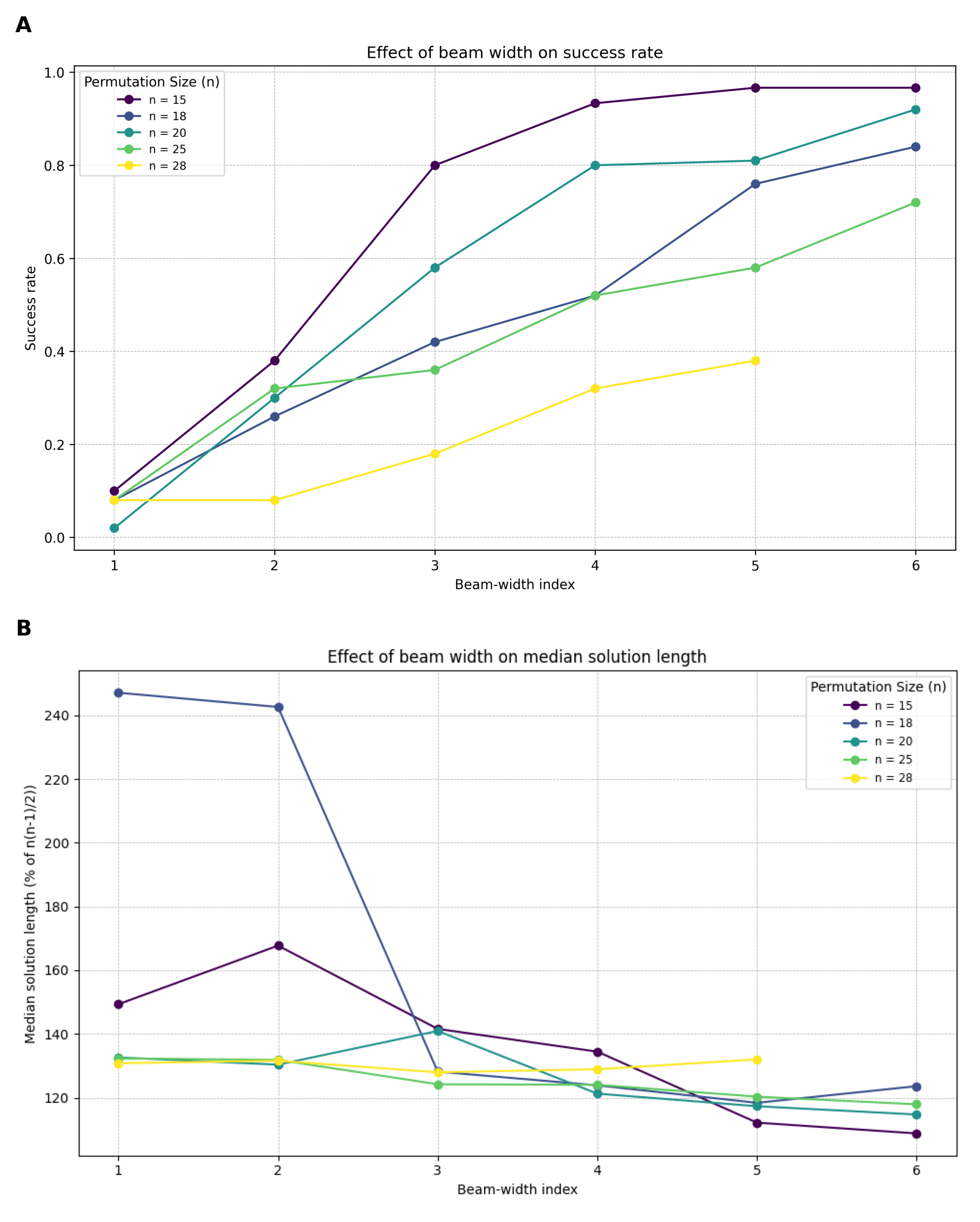}
  \caption{Effect of beam width on (A) success rate and (B) median solution length.
  Non-backtracking depth is fixed at two steps. Beam-width index $(i)$ is used
  instead of absolute beam widths: $i=1$ denotes the smallest beam width tested
  for each $n$, and each increment multiplies the beam width by two
  ($B_i = B_1 \cdot 2^{\,i-1}$).}
  \label{fig:bw_xgboost}
\end{figure}

To illustrate the impact of beam width on success rates, we used different beam widths at a fixed non-backtracking depth of two steps. Note that at step \(s\), we expand \(L_{s-1}\) to 1-move neighbors and discard any candidate whose hash appears in the stored layers (the last \(D\) frontier layers, where \(D\) is the non-backtracking depth). Since a 1-move neighbor
can only return to the immediate parent in \(L_{s-2}\), a 2-step non-backtracking depth (i.e., \(D=2\)) blocks exactly parent reversals; a 1-step depth (\(D=1\)) does not, because \(L_{s-2}\) lies outside the history. For each permutation size $n$ (15, 18, 20, 25, and 28), we started from the minimum beam width that yielded a non-zero success rate. Then, we increased it exponentially (in powers of two) 5 to 6 times. Since the minimal successful beam widths differ dramatically between small and large $n$ (e.g., 2 for $n = 15$ and 16384 for $n = 28$), we represented beam widths by their increment index $i$ rather than absolute values. Thus, $i=1$ always denotes the smallest beam width tested for each $n$, $i=2$ denotes the second smallest, and so forth. This indexing allowed us to plot success rates for all permutation sizes on a unified x-axis (Fig. \ref{fig:bw_xgboost}A).

This experiment clearly demonstrates a consistent trend within each $n$: increasing the beam width leads to higher success rates. However, the marginal gains diminish as $n$ increases. That is, larger permutations require significantly wider beams to achieve the same improvement in the success rate, indicating reduced search efficiency in higher-dimensional search spaces. Additionally, we show that wider beams somewhat reduce median solution lengths (Fig. \ref{fig:bw_xgboost}B), indicating more optimal paths.

\begin{figure}[H]
  \centering
  \includegraphics[width=0.82\textwidth]{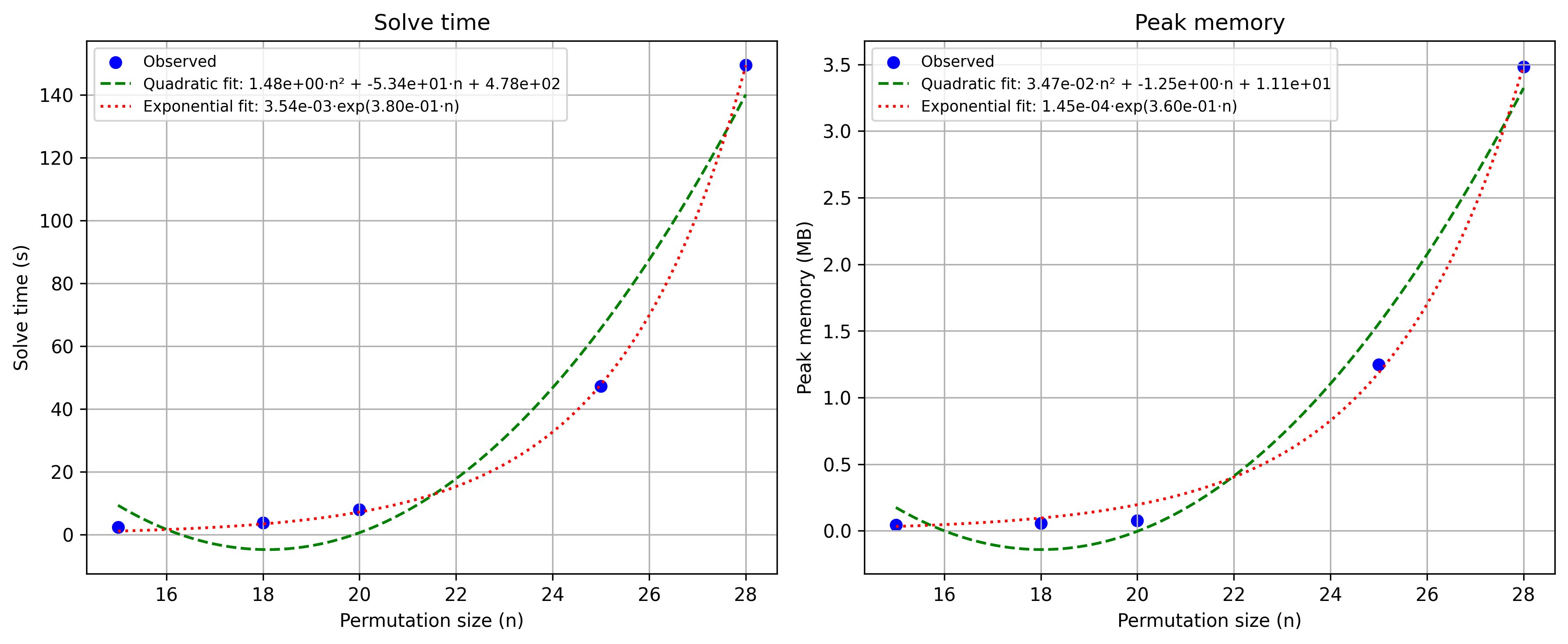}
  \caption{Solve time and peak memory as functions of permutation size $n$.
  Beam width and non-backtracking depth are fixed at high-performing values.
  The trends indicate approximately exponential growth with $n$ (complexity $\approx O(e^{n})$).}
  \label{fig:xgboost_time_mem}
\end{figure}

\enlargethispage{3em}
Finally, we assessed how computational resource demands scale with increasing permutation size when both beam width and non-backtracking depth are fixed at high-performing values. We found that both solution time and peak memory increase exponentially with $n$, revealing an asymptotic complexity of approximately $O(e^n)$ (Fig. \ref{fig:xgboost_time_mem}).

\subsection{Dependence on learning rate, epochs number and network size}
\label{sec-lr}
\begin{figure}[H]
  \centering
  \includegraphics[width=0.9\textwidth]{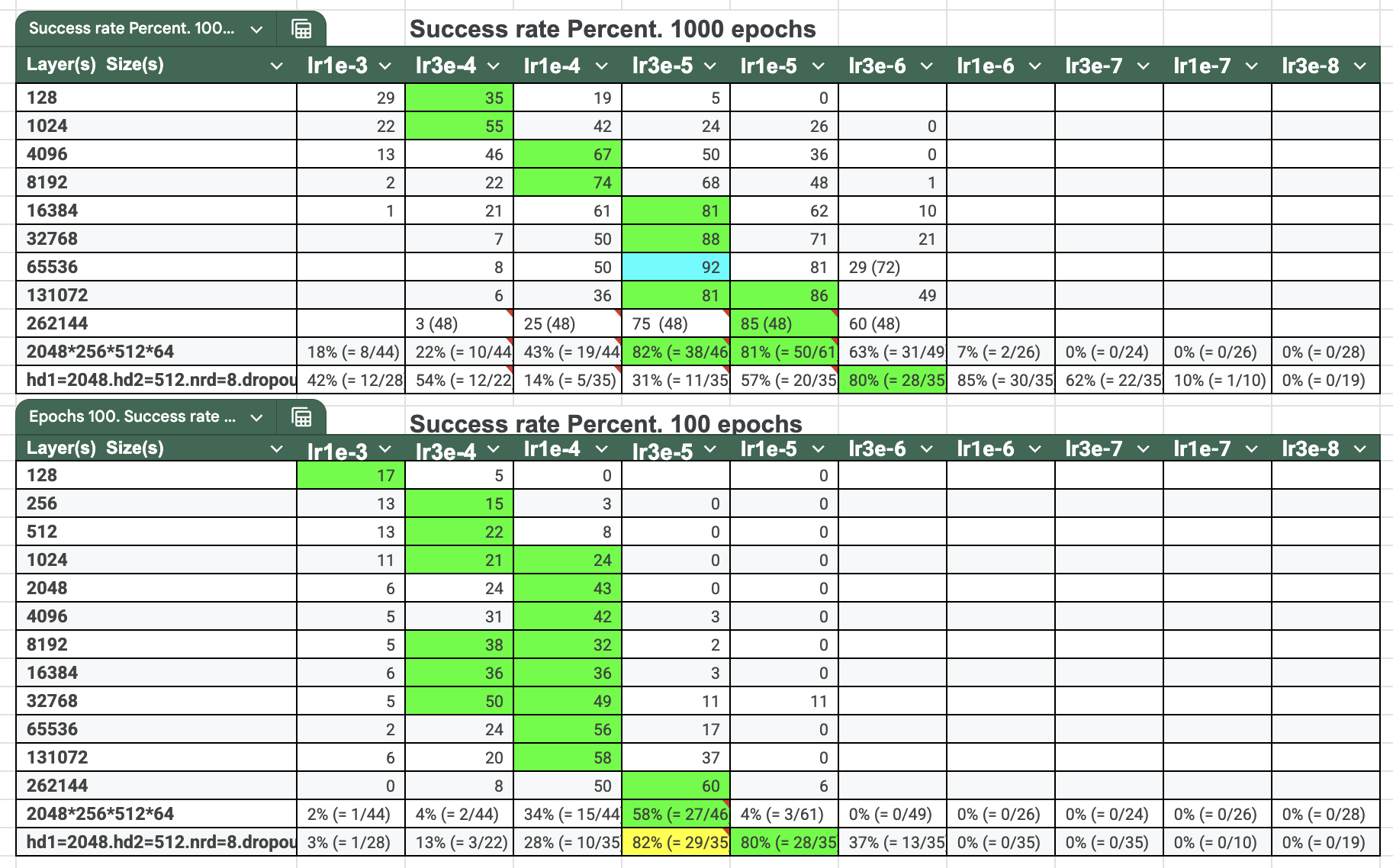}
  \caption{Success Rate for 4000 and 16000 epochs.}
  \label{fig:SuccessRate1}
\end{figure}

\begin{figure}[H]
  \centering
  \includegraphics[width=0.9\textwidth]{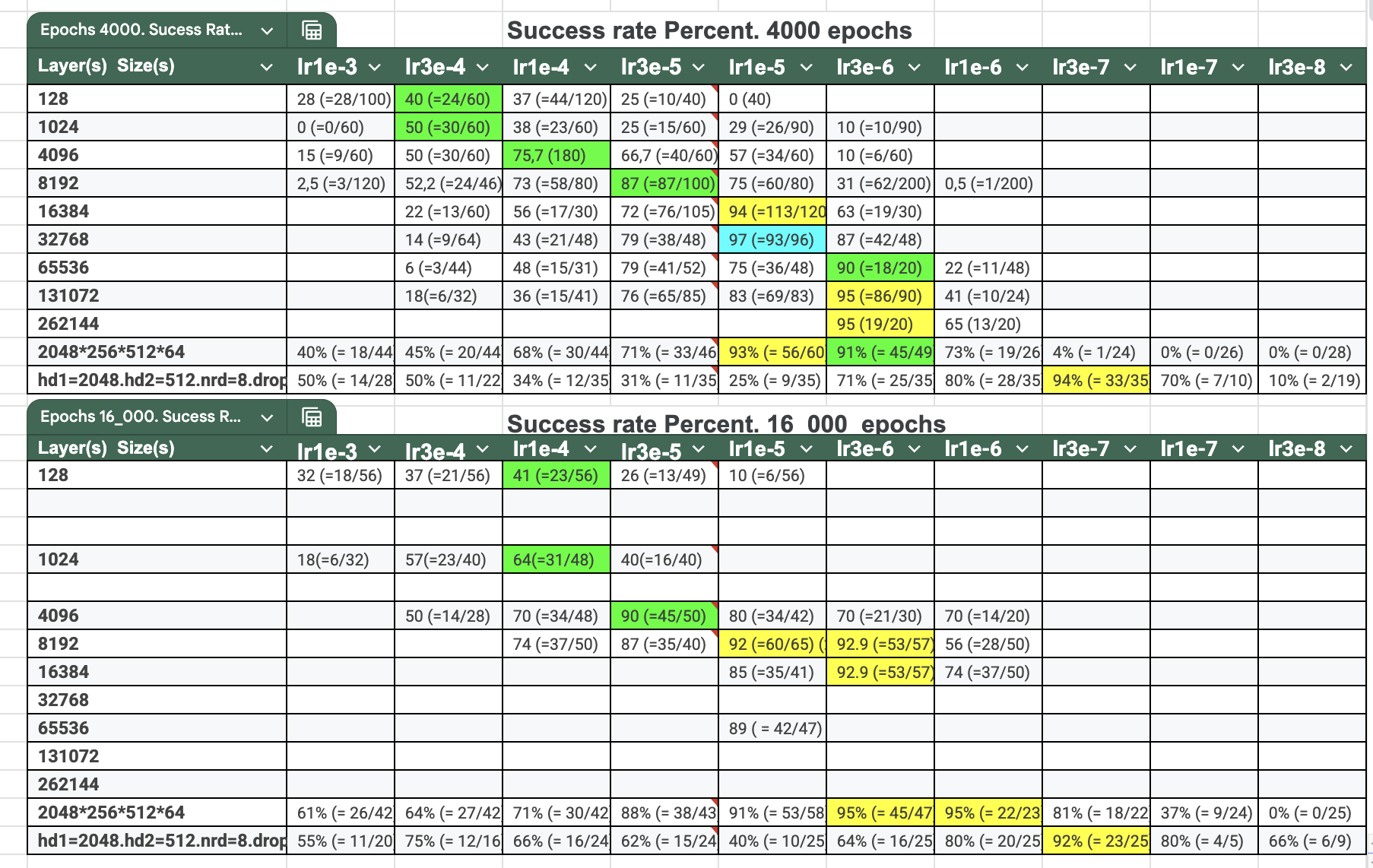}
  \caption{Success Rate for 4000 and 16000 epochs.}
  \label{fig:SuccessRate2}
\end{figure}

Here we present quite an interesting dependence on the combination of the  learning rate, epochs number and neural network size.
(More simulation results for various parameters can be found in \href{https://www.kaggle.com/competitions/lrx-oeis-a-186783-brainstorm-math-conjecture/discussion/612338}{spreadsheet here}, or  on \href{https://docs.google.com/spreadsheets/d/1KFYTqPDSdH8_vdvWz3YLTLKg5IlW1Kkx_9kJClSKGOs/edit?usp=sharing}{google sheets},
experiments are mostly done in the notebooks \href{https://www.kaggle.com/code/alexandervc/lrx-cayleypy-baseline2-mlp}{nptebook1}, \href{https://www.kaggle.com/code/alexandervc/lrx-cayleypy-rl-mdqn}{notebook2}.)
We mostly consider single layer MLP with layer size increasing by factor of 2 from 128 to 262144. 
Figures \ref{fig:SuccessRate1},\ref{fig:SuccessRate2} show success rates for $S_{16}$, beam size equal to 1,
pure diffusion distance training (without RL part). 
One can observe that, wider networks can achieve better results, however with smaller learning rate and larger number of epochs.
One also observe that optimal value of the learning rate is decreasing with number of epochs increased. 
Unfortunately we were unable to achieve 100\% success rate though results are quite close. 
Highlighted cells show best results in a row, blue highlight best result for table, yellow - if results are insignificantly different from the best, "lr" is learning rate. Number of experiments is typically 50-100, or shown in the brackets.
Typically one stops learning by when the validation loss stops to decrease, but in that case we observe that this standard strategy does not work, validation loss stabilizes quite early, however additional training improves the actual solution ability.
\enlargethispage{3em}
We attribute that to the noisy character of random walk data - loss cannot be made in zero in that situation in principle,
because each training batch (random walks) is different from the perfect diffusion distance. 
Thus we need to understand how many epochs to train for the given learning rate. 

\subsection{A Tropical Perspective on the Bellman Equation}\label{sec-tropical}
Tropical mathematics studies structures defined over a semiring equipped with the minimum and addition operations, instead of the usual addition and multiplication. The Bellman equation naturally embodies the tropical spirit due to the presence of the minimum over neighbors, in contrast to the averaging operations more common in graph theory. Here we discuss some elementary statements clarifying this connection.

Consider a weighted directed graph $G$ with $n$ vertices. Its tropical adjacency matrix (\cite{MacSturm}, Chapter~1, Section~1.2) $A_{G},$ where the off-diagonal elements represent the lengths of the edges $a_{ij}\in\mathbf{R},$ and the diagonal elements are set to $a_{ii}=0.$ If there is no edge $(i, j)$ in $G,$ the corresponding entry in the matrix is assigned $a_{ij}=+\infty.$

{\bf Tropical $(n-1)$-power - lengths of shortest paths.}
It is easy to see (e.g. proposition~1.2.1 (\cite{MacSturm}, Chapter~1, Section~1.2) the tropical multiplication of the adjacency matrix $A_{G}$ with itself $(n - 1)$ times results in the matrix $P_{G}$ where the element $p_{ij}$ represents the lengths shortest path from the vertex $i$ to the vertex $j$ in $G.$ 

That is similar to the well-known fact, that matrix elements of standard $k$-powers of adjacency matrices give {\bf numbers} of paths of length $k$, but tropical powers give {\bf lengths} of paths, not their counting. 

Let $G$ be a connected graph with vertex set $\{v_{i}\}$ and edges of unit lengths.
Then the $k$-th row of the matrix $P_{G}$ satisfies the Bellman equation (\ref{bel_eq}) with a boundary condition $d(v_{k}) = 0.$ In fact, the $(k,j)$ entry of this row equals the distance of the shortest path from $v_{k}$ to  $v_{j}$.

Notice that if the graph $G$ is connected, the matrix $P_{G}$ is represented by a weighted directed graph $G(P),$ that has a directed path from any vertex to any other vertex, which means that it is strongly connected. Then, Theorem~5.1.1 (\cite{MacSturm}, Chapter~5, Section~5.1) holds, stating that the matrix $P_{G}$ has only one eigenvalue, which is equal to 0.

Next, applying Theorem 5.1.3 (\cite{MacSturm}, Chapter~5, Section~5.1), we obtain that the columns of the matrix $P_{G}$ belong to the eigenspace of the tropical adjacency matrix $A_{G}.$ Thus, the solution to the Bellman equation can be associated with the problem of finding the eigenvectors of the matrix $A_{G}$ of the graph $G$ in tropical algebra.

Overall, one can think of the Bellman equation as a tropical analogue of the graph Laplacian, a central object of study in graph theory.
An intriguing question is: to what extent do the rich theories surrounding the Laplacian admit tropical counterparts?
In particular, since the cokernel of the Laplacian defines the sandpile (or Jacobian) group of a graph, one may ask whether there exists a tropical analogue of this construction.



\address{Institut Curie\\
Paris, 94260, France\\
\email{al.chervov@gmail.com}}

\address{IHES\\
 Paris, 91440, France\\
\email{asoibelman@gmail.com}}

\address{Kazakh-British Technical University\\
Almaty, 050000, Kazakhstan\\
\email{smlytkin@gmail.com}}

\address{Accenture\\
Mountain View, CA 94040, USA\\
\email{igor.kiselev@accenture.com}}

\address{Yandex\\
Saint Petersburg, 195027, Russia\\
\email{ifserge@gmail.com}}

\address{Meta\\
London, NW1 3FG, UK\\
\email{andlukyane@gmail.com}}

\address{Independent Researcher\\
Moscow, 117208, Russia\\
\email{an.dolgorukova@gmail.com}}

\address{Institute of Molecular Biology and Genetics\\
Kiev, 03143, Ukraine\\
\email{ogurtsov.a.b@gmail.com}}

\newpage
\address{Department of Mathematics and Computer Science, Saint Petersburg State University
Saint Petersburg, 199178, Russia\\
\email{fedyapetrov@gmail.com}}

\address{Stanford University\\
Stanford, CA 94305, United States\\
\email{krymskiy.stas@gmail.com}}

\address{University of Bologna\\
 Bologna, 40126, Italy\\
\email{mixnota@gmail.com}}

\address{Sobolev Institute of Mathematics\\
 Novosibirsk, 630090, Russia\\
\email{mathmanlily@gmail.com}}

\address{University of Toronto\\
 Toronto, M5S2E4, Canada\\
\email{denis.gorod@gmail.com}}

\address{Moscow State University\\
 Moscow, 119991, Russia\\
\email{grigoriy.rus@gmail.com}}

\address{Moscow State University\\
 Moscow, 119991, Russia\\
\email{scigverbii@gmail.com}}

\address{Independent Researcher\\
Amsterdam, 1098 , Netherlands\\
\email{zamkovoyvladislav@gmail.com}}

\address{MIREA-Russian technological university\\
Moscow, 119454, Russia\\
\email{liuda.tarusina@gmail.com}}

\address{MIREA-Russian technological university\\
Moscow, 119454, Russia\\
\email{ivankolt@gmail.com}}
\newpage
\address{MIREA-Russian technological university\\
Moscow, 119454, Russia\\
\email{arseniysychev2015@gmail.com}}

\address{Research Center of the Artificial Intelligence Institute, Innopolis \\
Innopolis, 420500, Russia\\
\email{obozovmark9@gmail.com, author: Mark Obozov}}

\address{Yandex\\
Moscow, 119021, Russia\\
\email{greatrobocreator@gmail.com}}

\address{Independent Researcher\\
Moscow, 117105, Russia\\
\email{nikolenko.sergei@icloud.com}}

\address{International University of Kyrgyzstan\\
Bishkek, 720001, Kyrgyzstan\\
\email{nnurseultan07@gmail.com}}

\address{Independent Researcher\\
Astana, 010000, Kazakhstan\\
\email{rustem.turtayev@alumni.nu.edu.kz}}

\address{Incarta\\
San Francisco, CA 94103, USA\\
\email{contact@rokotyan.com}}

\address{Yandex\\
Belgrade, 11000, Serbia\\
\email{iggisv9t@gmail.com}}

\address{Independent Researcher\\
Moscow, 117638, Russia\\
\email{thehdotx@gmail.com}}

\address{Independent Researcher\\
Kharkov, 61022, Ukraine\\
\email{veniamin.nelin@gmail.com}}
\newpage
\address{Independent Researcher\\
Moscow, 125167, Russia\\
\email{ermilov\_s@mail.ru}}

\address{Independent Researcher\\
Nizhnii Novgorod, 603950, Russia\\
\email{lido4ka.sh@mail.ru}}

\address{Independent Researcher\\
Astana, 010000, Kazakhstan\\
\email{dmamayeva367@gmail.com}}

\address{Peoples' Friendship University of Russia (RUDN University)\\
Moscow, 117198, Russia\\
\email{t0nykar@yandex.ru}}

\address{Technical University of Munich\\
Munich, 80333, Germany\\
\email{khoruzhii.ka@gmail.com}}

\address{Institute of artificial intelligence, MIREA-Russian technological university\\
Moscow, 119454, Russia\\
\email{romanov@mirea.ru}\\
\received{February 27, 2025}\\
\accepted{August 09, 2025}}
\end{document}